\documentclass[10pt,journal,cspaper,compsoc]{IEEEtran}
\usepackage{epsfig}
\usepackage{graphicx}
\usepackage{amsmath}
\usepackage{amssymb}
\usepackage{epstopdf}
\usepackage{algorithm}
\usepackage{algorithmic}
\usepackage{subfigure}

\ifCLASSOPTIONcompsoc
\else
\fi
\ifCLASSINFOpdf
\else
\fi
\usepackage{url}

\begin{document}
%
\title{SIRF: Simultaneous Image Registration and Fusion in A Unified Framework}
%
%
%
%

\author{Chen~Chen,
        ~Yeqing~Li,
        ~Wei~Liu, 
        and~Junzhou~Huang*
\IEEEcompsocitemizethanks{\IEEEcompsocthanksitem C. Chen, Y. Li and J. Huang are with the Department of Computer Science and Engineering, University of Texas at Arlington, Texas 76019, USA.
Corresponding author: Junzhou Huang. Email: jzhuang@uta.edu.
\IEEEcompsocthanksitem W. Liu is with IBM T.
J. Watson Research Center.}
\thanks{}}

%
%

\markboth{JOURNAL OF LATEX CLASS FILES}%
{Shell \MakeLowercase{\textit{et al.}}: Bare Demo of IEEEtran.cls for Computer Society Journals}
%


\IEEEcompsoctitleabstractindextext{%
\begin{abstract}

  In this paper, we propose a novel method for image fusion with a high-resolution panchromatic image and a low-resolution multispectral image at the same geographical location.
  The fusion is formulated as a convex optimization problem which minimizes a linear combination of a least-squares fitting term and a dynamic gradient sparsity regularizer.
  The former is to preserve accurate spectral information of the multispectral image, while the latter is to keep sharp edges of the high-resolution panchromatic image.
  We further propose to simultaneously register the two images during the fusing process, which is naturally achieved by virtue of the dynamic gradient sparsity property. An efficient algorithm is then devised to solve the optimization problem, accomplishing a linear computational complexity in the size of the output image in each iteration.
  We compare our method against seven state-of-the-art image fusion methods on multispectral image datasets from four satellites. Extensive experimental results demonstrate that the proposed method substantially outperforms the others in terms of both spatial and spectral qualities. We also show that our method can provide high-quality products from coarsely registered real-world datasets. Finally, a MATLAB implementation is provided to facilitate future research.

\end{abstract}

}

\maketitle

\IEEEdisplaynotcompsoctitleabstractindextext

%
\IEEEpeerreviewmaketitle

\section{Introduction}

\begin{figure}[htbp]
\centering \vspace{-0.0cm}
        \includegraphics[scale=0.5]{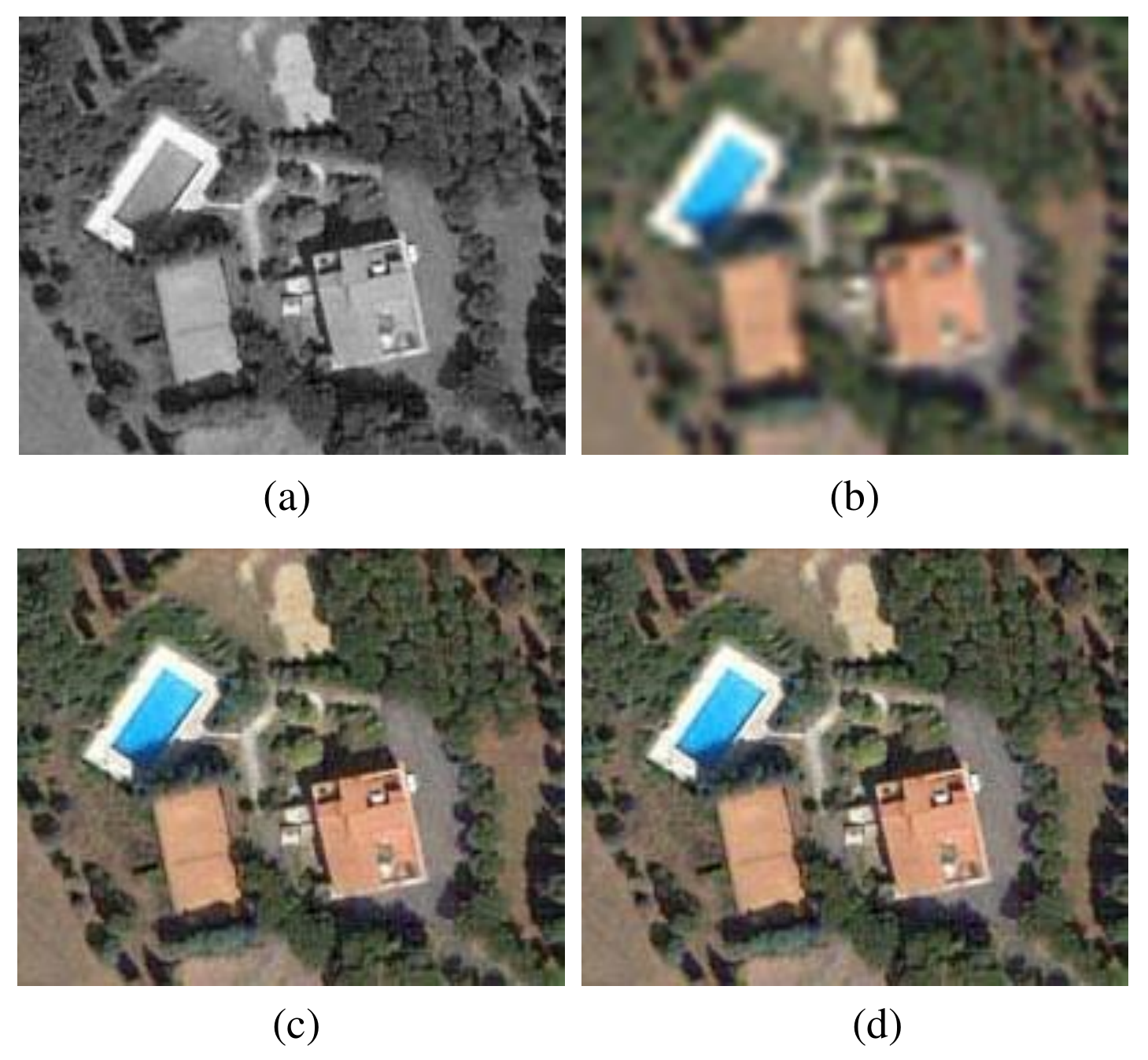}
\caption{(a) A high-resolution panchromatic image. (b) The corresponding low-resolution multispectral image. (c) Our fusion result. (d) The ground-truth. 
}
\label{fig:qb}
\end{figure}

\IEEEPARstart{M}ultispectral (MS) images are widely used in many fields of remote sensing such as environmental monitoring, agriculture, mineral exploration, \textit{etc}. However, the design of MS sensors with high resolution is confined by infrastructure limits in onboard storage and bandwidth transmission \cite{thomas2008synthesis}. In contrast,
panchromatic (Pan) gray-scaled images with high spatial resolution can be obtained more conveniently, because they are composed of much reduced numbers of pixels. The combinations of Pan images in high spatial resolution and MS images in high spectral resolution can be acquired simultaneously from most existing satellites. Therefore, we expect to obtain images with both high spatial and high spectral resolutions via image fusion, which is also called pan-sharpening in the literature. A fusion example on Quickbird satellite images is shown in Figure \ref{fig:qb}.

\vspace{-0.0cm}
 \vspace{-0.0cm}

Image fusion is a typical inverse problem and generally difficult to handle. The first question is how to preserve accurate information from both the Pan and MS images.
A number of conventional methods use projection and substitution, which include principal component analysis (PCA) \cite{chavez1991comparison}, intensity hue saturation (IHS) \cite{haydn1982application}, wavelet \cite{zhou1998wavelet}, and their combinations. These methods perform fusion in the following scheme: upsampling, forward transform, intensity matching, component substitution, and reverse transform \cite{amro2011survey}. Other methods such as Brovey \cite{gillespie1986color} assume that the Pan image is a linear combination of all bands of the fused image. A detailed survey of the existing methods can be found in \cite{amro2011survey}.
While previous classical methods provided some good visual results, they are very likely to suffer from spectral distortion. The reason is that
they all made strong assumptions that are not realistic from the viewpoint of remote sensing physics \cite{thomas2008synthesis}.

In order to overcome the issue caused by spectral distortion, a suite of variational approaches have emerged recently,
all of which formulate an energy function based on somewhat weak assumptions, and then optimize such a function to obtain the optimum.
These methods are also called model-based fusion \cite{aanaes2008model}. The earliest variational method P+XS \cite{ballester2006variational}
is based on the linear combination assumption used by Brovey \cite{gillespie1986color}, and additionally assumes that the upsampled MS image
is the fused result after blurring. As an accurate blur kernel is difficult to pre-estimate, AVWP \cite{moller2012variational} replaces this
term with a spectral ratio constraint to preserve spectral information, and meanwhile forces the fused image close to the wavelet fused image \cite{zhou1998wavelet}. Another variational model is engaged in estimating the fused image in conjunction with the blurring model parameters iteratively \cite{fang2013variational}.
Owing to the success of compressive sensing \cite{candes2006robust}, some methods are proposed to employ sparsity regularization and dictionary learning
to tackle image fusion \cite{ding2014pan,li2011new,jiang2012practical,palsson2014new,li2009fusion}. In the recent study \cite{aly2014regularized},
a highpass filter is introduced to compensate the lowpass filter for spectral observation.
Promising results achieved by these variational methods have shown that they can reduce spectral distortion.
However, due to the lack of an effective model to preserve spatial information, visible artifacts or blurriness
may appear in the fused results. Moreover, all these methods often involve high computational complexities
which prevent these methods from being scalable to massive datasets.

The second question in fusion is how to reduce the effect of misalignments.
Almost all the above methods require a precise registration before fusion.
However, pre-registration is quite challenging due to the significant resolution difference between input images \cite{inglada2007analysis,leprince2007automatic,uss2014precise}. After pre-registration, 0.5-pixel's misalignment on the multi-spectral image corresponds to 2-pixels's misalignment on the Pan image when the resolution difference is four times.
It has been shown that a geometrical distortion of only 0.1 pixel in standard deviation yields a significant impact on pixel-to-pixel fusion \cite{blanc1998importance}.
The existing fusion methods have to ``tolerate" such misalignments. Therefore, fusion accuracy is inevitably compromised for real-world datasets.

In this paper, we propose a novel method for \textit{Simultaneous Image Registration and Fusion}, named SIRF by the initials.
We assume that the fused image after downsampling should be close to the input MS image, which is formulated into a least-squares fitting term to keep spectral information.
Motivated by the geographical relationship between the fused image and the input Pan image, a dynamic gradient sparsity property is discovered, defined, and then exploited
to improve spatial quality. Importantly, we find that the combined model does not violate remote sensing physics,
and that the dynamic gradient sparsity naturally induces accurate registration even under severe intensity distortions.
Moreover, our method incorporates the inherent correlation of different bands, which has been rarely considered before.
To optimize the entire energy function, a new algorithm is designed based on the proximal gradient technique.
In specific, we solve the subproblems efficiently by applying the fast iterative shrinkage-thresholding algorithm (FISTA) \cite{beck2009afast}
and the gradient descent method with backtracking, respectively. The overall algorithm remains a linear computational complexity in each iteration,
and is thus scalable to massive datasets. The algorithm can directly be applied to real-world datasets without prefiltering, feature extraction, training, \textit{etc}.
Finally, there is only one non-sensitive parameter in SIRF, which is another advantage comparing with existing variational methods.
Extensive experimental results demonstrate that our method can significantly reduce spectral distortions while preserving sharp object
boundaries in fused images. In particular, our method is shown to be much more powerful than the competing methods
on real-world datasets with pre-registration errors.


The rest of this paper is organized as follows. We define the dynamic gradient sparsity and give the formulation of SIRF in Section \ref{sec:mod}.
In Section \ref{sec:alg}, an efficient algorithm is presented to solve the optimization problem. The experimental results are shown in Section \ref{sec:exp}.
Finally, we conclude the paper in Section \ref{sec:con}. Some preliminary results of this work have been presented in our prior paper \cite{chen2014image}.

\section{Modeling} \label{sec:mod}

\subsection{Notations}
Scalers are denoted by lowercase letters. Bold letters
denote matrices. Specially, $\textbf{P} \in \mathbb{R}^{m \times n}$
denotes the Pan image and $\textbf{M} \in \mathbb{R}^{\frac{m}{c}
\times \frac{n}{c} \times s}$ denotes the low-resolution MS image.
$c$ is a constant. For example $c=4$ when the resolution of Pan
image is 0.6m and that of MS image is 2.4m in Quickbird acquisition.
The image to be fused is denoted by
$\textbf{X} \in \mathbb{R}^{m \times n \times s}$. 
$||\cdot||_F$ denotes the Frobenius norm. For simpleness,
$\textbf{X}_{i,j,d}$ denotes the element in $i$-th row, $j$-th
column and $d$-th band in \textbf{X}. And $\textbf{X}_d$ denotes the
whole $d$-th band, which is therefore a matrix.

\subsection{Local Spectral Consistency}

Many existing methods upsample the MS image and extract spectral information from this upsampled MS image. However, the upsampled image is blurry and not accurate.
Therefore, we only assume the fused image after downsampling is close to the {original} MS image.  Least squares fitting is used to model this relationship:
\begin{eqnarray}
E_1 = \frac{1}{2} || \psi \textbf{X} - \textbf{M} ||_F^2,
 \label{eqn:jtvfusion}
\end{eqnarray}
where $\psi$ denotes a downsampling operator. Local spectral
information is forced to be consistent with each MS pixels. 
This function is physically motivated and thus can avoid spectral distortion in the result.

Minimizing $E_1$ would be a severely ill-posed problem, due to the very low undersampling rate (\emph{e.g.}, 1/16 when $c=4$). Without strong prior information, $\textbf{X}$ is almost impossible to be estimated accurately.

\subsection{Dynamic Gradient Sparsity}


Fortunately, the Pan image provides such prior information. Due to
the strong geographical correlation with the fused image
$\textbf{X}$, the Pan image has already provided us with clear
edge information of land objects. Many researchers
attempt to build this relationship mathematically. From recent
reviews \cite{thomas2008synthesis,amro2011survey}, however, few methods can effectively characterize this relationship.

As remotely sensed images are often piece-wise smooth, their gradients tend to be sparse and the non-zeros corresponds to the edges. In addition, the positions of such edges should be the same as those on the Pan image when the images have been well aligned. It demonstrates that the sparsity property is not fixed but dynamic according to a reference image. This property has not been studied in sparsity theories yet. We call the data with such a property a dynamic gradient sparse signal/image.


\vspace{0.1cm}
\textbf{Definition}: \emph{Let $x \in \mathbb{R}^N$ and $r\in \mathbb{R}^N$ denote the signal and the reference signal. $\Omega_x$ and $\Omega_r$ denote the support sets\footnote{Here we mean the indices of the non-zero components.} of their gradients, respectively. The set of dynamic gradient sparse signals is defined as:} \\
$\mathcal{S}_x = \{ x \in \mathbb{R}^N: |\Omega_x| = K, \Omega_x = \Omega_r, \ \mathrm{with} \ K<<N  \}$. \vspace{0.3cm}

Using similar logic, it can be extended to multi-channel/spectral signals
and images. The first term in P+XS \cite{ballester2006variational}
and AVWP \cite{moller2012variational} does not induce sparseness and
tends to over-smooth the image by penalizing large values. In
\cite{fang2013variational}, the first term is derived from the
linear combination assumption in P+XS; it does not promote sparsity
for each band. In \cite{palsson2014new}, total variation is used to encourage sparseness of the gradients. However, no reference information from the Pan image is integrated in this regularization term.  Different from previous work, dynamic gradient
sparsity is encouraged in our method.
Beside the prior information
that previous methods attempt to use, we also notice the intra-
correlations across different bands as they are the representations
of the same land objects. Therefore, the gradients of different
bands should be group sparse. It is widely known that the $\ell_1$ norm encourages
sparsity and the $\ell_{2,1}$ norm encourages group sparsity
\cite{yuan2005model}. Thus we propose a new energy function to
encourage dynamic gradient sparsity and group sparsity
simultaneously:
\begin{eqnarray}
E_2 =&||\nabla{\textbf{X}} - \nabla {D}(\textbf{P})||_{2,1} \\
=&\sum_{i}\sum_{j}\sqrt{\sum_{d} \sum_q
(\nabla_{q}\textbf{X}_{i,j,d} - \nabla_{q}\textbf{P}_{i,j})^2 },
 \label{eqn:jtvfusion}
\end{eqnarray}
where $q={1,2}$ and $\emph{D}(\textbf{P})$ means duplicating
$\textbf{P}$ to $s$ bands. Interestingly, when there is no reference
image, \emph{i.e.}, \textbf{P} = \textbf{0}, the result is
identical to that of vectorial total variation (VTV)
\cite{bresson2008fast,huang2012fast,chen2013calibrationless}, which is widely used in multi-channel image
denoising, deblurring and reconstruction.


\vspace{-0.0cm}
\begin{figure}[htbp]
\centering \vspace{-0.0cm}
        \includegraphics[scale=0.3]{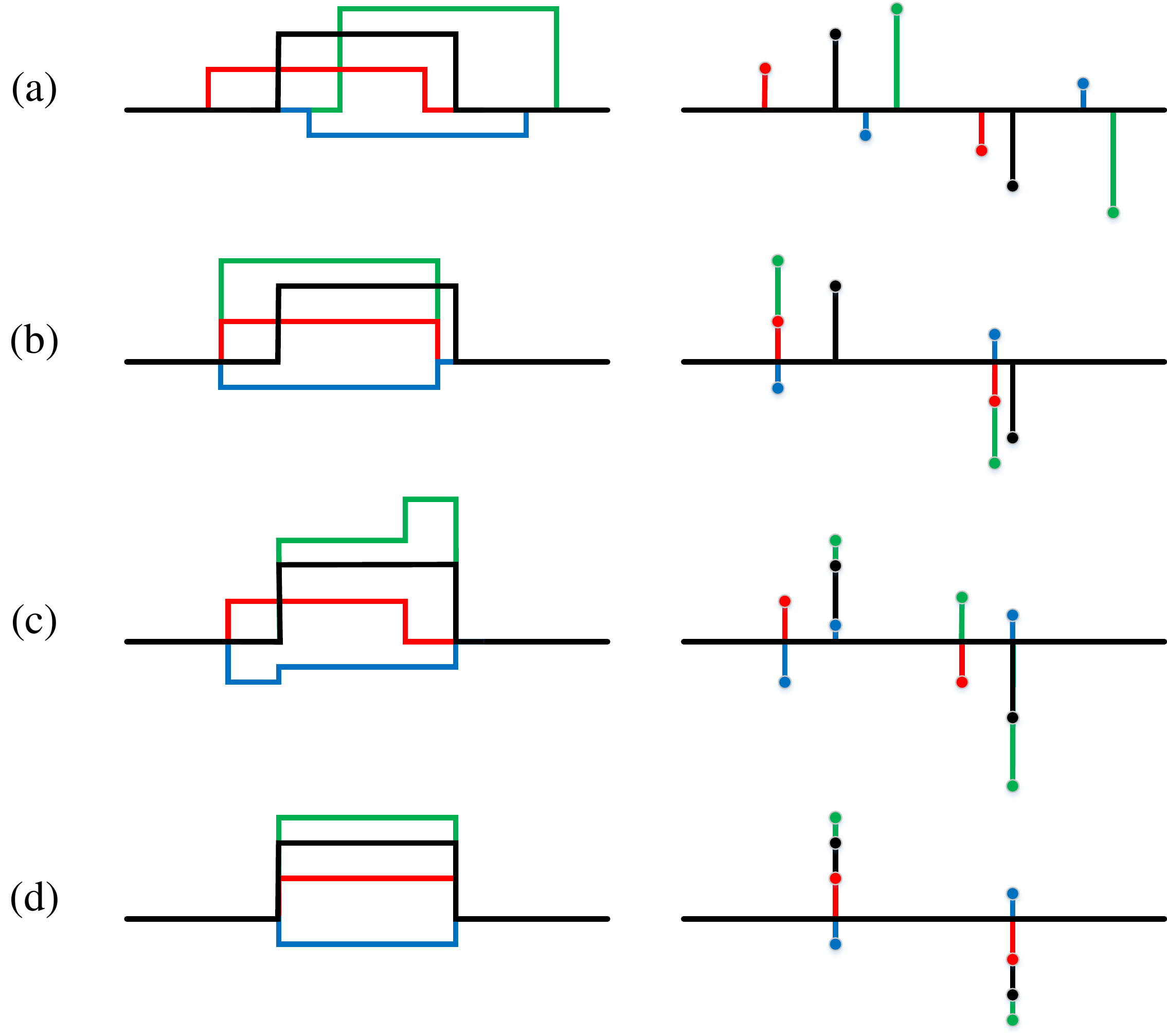}
\caption{ Illustration of possible solutions for different gradient based penalties. The black denotes a reference signal. RGB color lines denotes the solutions of different models. Left: 1D signals. Right: the corresponding gradients. (a) A possible solution of TV \cite{palsson2014new}: the gradients of RGB channels are sparse but may not be correlated. (b) A possible solution of VTV: the gradients of R, G, B channels are group sparse, but may not be correlated to the reference signal. (c) A possible solution of \cite{fang2013variational}:  it does not encourage sparseness for each channel individually. (d) A possible solution of dynamic gradient sparsity regularization: the gradients can only be group sparse following the reference.}
\label{fig:dgs}\vspace{-0.0cm}
\end{figure} \vspace{-0.0cm}




To demonstrate why $E_2$ encourages dynamic gradient sparsity, we show a simple example on a 1D multi-channel signal in Figure \ref{fig:dgs}. 
We could observe that, if the solution has a different support set
from the reference, the total sparsity of the gradients will be
increased. Cases (a)-(d) have group sparsity number 8, 4, 4, 2
respectively. Therefore, (a)-(c) will be penalized because they are
not the sparsest solution in our method.

Combining the two energy functions, the image fusion problem can be formulated as:
\begin{flalign}
\min_{\textbf{X}} \{ &E(\textbf{X}) = E_1 + \lambda E_2 \notag \\
&= \frac{1}{2} || \psi \textbf{X} - \textbf{M}||_F^2 + \lambda||\nabla{\textbf{X}} - \nabla {D}(\textbf{P})||_{2,1} \}, \label{eqn:jtvfusion}
\end{flalign}
where $\lambda$ is a positive parameter.

\begin{figure*}[htbp]
\centering \vspace{-0.0cm}
        \includegraphics[scale=0.4]{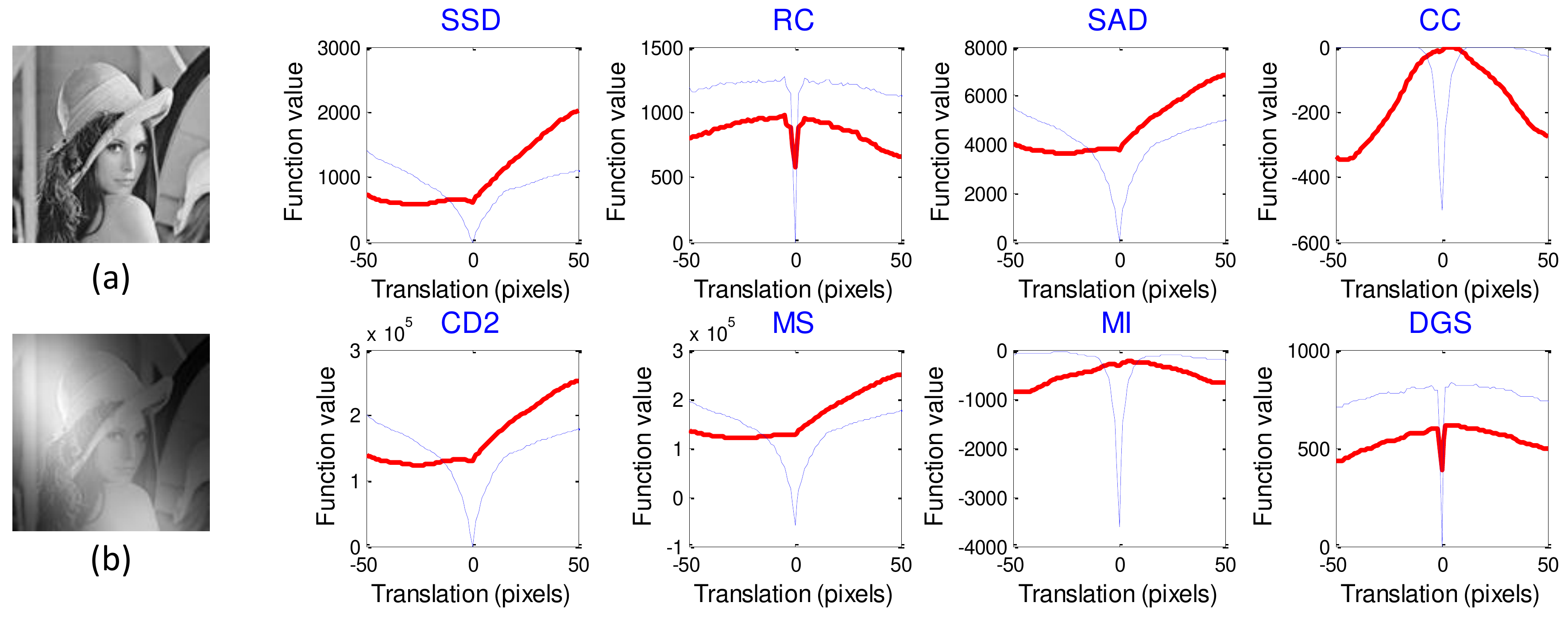}
\caption{A toy registration example with respect to horizontal translation using different similarity measures (SSD \cite{thevenaz1998pyramid}, RC \cite{myronenko2010intensity}, SAD, CC, CD2  \cite{cohen2002new}, MS \cite{myronenko2009maximum}, MI \cite{viola1997alignment} and the proposed dynamic gradient sparsity (DGS)). (a) The Lena image ($128 \times 128$). (b) A toy Lena image under a severe intensity distortion. Blue curves: registration between (a) and (a); red curves: registration between (b) and (a).}
\label{fig:ss}
\end{figure*}
Comparing our method with existing methods,
the first benefit of our method comes from the local spectral
constraint. It does not rely on the upsampled MS image and
linear-combination assumption. Therefore, only accurate spectral
information is kept. Second, the proposed
dynamic gradient sparsity only forces the support sets to be the
same, while the signs of the gradients as well as the magnitudes of
the signal are not required to be the same. These properties
make it invariant under contrast inversion
\cite{thomas2008synthesis} and not sensitive to illumination
conditions. It is possible to be applied to image fusion from
different sources or at different acquisition times. Last but not least, only our method can jointly fuse
multiple bands simultaneously, which provides robustness to noise \cite{huang2011learning,huang2009learningb}.
These advantages exist in our method uniquely.

%

%

\subsection{Simultaneous image registration}

The image fusion process (\ref{eqn:jtvfusion}) requires accurate registration between the Pan image and the multispectral image, while the misalignment is very difficult to eliminate during preprocessing.
To overcome this difficulty, we propose to register the images simultaneously during the fusion process. On one hand, the multispectral image is sharpened to higher resolution. This allows us to register the images more accurately. On the other side, the misalignment is gradually eliminated and the image can be fused more precisely. We iteratively run these two processes until convergence.

In the literature, existing image registration methods can be classified into feature-based registration (\emph{e.g.}, \cite{zheng2013landmark,jiayi2014robust}) and pixel-based (or intensity-based) registration (\emph{e.g.}, \cite{myronenko2010intensity,peng2012rasl}), based on the features used in registration. Feature-based methods rely on the landmarks extracted from the images. However, locating reliable features is still an open problem and an active topic of research \cite{sotiras2013deformable}. 
Here, we are more interested in intensity-based registration, which can be combined with fusion in a unified optimization scheme.

One of the most essential components of image registration is the energy function to measure (dis)similarity. The optimized similarity should correspond to the correct spatial alignment.
There are a few similarity measures have been used for registration, to name a few,
sum-of-squared-difference (SSD) \cite{thevenaz1998pyramid}, residual complexity (RC) \cite{myronenko2010intensity}, sum-of-absolute value (SAD), correlation coefficient (CC),  CD2 \cite{cohen2002new},
MS \cite{myronenko2009maximum} and mutual information (MI) \cite{viola1997alignment,suri2010mutual}. When the input images have similar intensity fields, all of existing similarity measures are able to find the true transformation based on the pixel values. However, due to the physics of remote sensing \cite{thomas2008synthesis}, the intensity fields of remotely sensed images may vary significantly (\emph{e.g.}, aquired under different illumination conditions, or by different types of sensors). Lots of existing intensity-based similarity measures are not robust to such intensity distortions, \emph{e.g.}, the widely used SSD. Although some methods are proposed for simultaneous registration and intensity correction \cite{pohl2005unifying,modersitzki2006combining}, they often involve much higher computation complexity and multiple local minima. Considering the large image size in remote sensing, we need a stable similarity measure without introducing high computational complexity.

Fortunately, we already have one. We use the dynamic gradient sparsity to preserve spatial information (Fig. \ref{fig:dgs}). Any misalignment will increase the sparsity of the gradients. Thus, the dynamic gradient sparsity can be naturally used as a similarity measure. We can revise the energy function for simultaneous image registration and fusion:

\begin{flalign}
E(\textbf{X},\mathcal{T}) = \frac{1}{2} || \psi \textbf{X} - \textbf{M}||_F^2 + \lambda||\nabla{\textbf{X}} - \nabla \mathcal{T}({D}(\textbf{P}))||_{2,1}, \label{eqn:sirf}
\end{flalign}
where $\mathcal{T}$ is the transformation to be estimated.

We compare the proposed similarity measure with existing approaches in Fig. \ref{fig:ss}, with SSD \cite{thevenaz1998pyramid}, RC \cite{myronenko2010intensity}, SAD, CC, CD2  \cite{cohen2002new}, MS \cite{myronenko2009maximum} and MI \cite{viola1997alignment}. 
The Lena image is registered with itself with respect to the horizontal translations.
The blue curves in Fig. \ref{fig:ss} show the responses of different measures, all of which can find the optimal alignment at the zero translation. After adding intensity distortions and rescaling, the appearance of source image shown in Fig. \ref{fig:ss}(b) is not consistent with that of the original Lena image. The results denoted by the red curves show that only RC and the proposed method can handle this intensity distortion while all the other methods fail. In RC, the discrete cosine transform (DCT) is used to sparisify the residual image, which is has $\mathcal{O}(N\log N)$ complexity. Our method only has linear complexity in each iteration. More importantly, it is unknown how to combine RC in the image fusion process, while we can simultaneously achieve image registration and fusion in the unified model (\ref{eqn:sirf}).

\section{Algorithm} \label{sec:alg}

Now our goal is to minimize the energy function (\ref{eqn:sirf}).
We first solve the problem with respect to $\textbf{X}$ by fixing $\mathcal{T}$, and then solve the problem with respect to $\mathcal{T}$ by fixing $\textbf{X}$.
For the $\textbf{X}$ subproblem:
\begin{flalign}
E(\textbf{X}) = \frac{1}{2} || \psi \textbf{X} - \textbf{M}||_F^2 + \lambda||\nabla{\textbf{X}} - \nabla \mathcal{T}({D}(\textbf{P}))||_{2,1}, \label{eqn:sub1}
\end{flalign}
it is a obvious convex function. The first term is smooth while the second term is non-smooth. This motivates us to solve this subproblem in the FISTA framework \cite{beck2009afast,huang2011composite}. It has been proven that FISTA can achieve the optimal convergence rate for first order methods. That is, $E(\textbf{X}^k)-E(\textbf{X}^*) \sim \mathcal{O}(1/k^2)$, where $\textbf{X}^*$ is the optimal solution and $k$ is the iteration counter.

The second subproblem with respect to $\mathcal{T}$ can be written as:
\begin{flalign}
\min E(\mathcal{T}) = ||\nabla{\textbf{X}} - \nabla \mathcal{T}({D}(\textbf{P}))||_{2,1}. \label{eqn:sub2}
\end{flalign}
We solve this subproblem during each iteration of FISTA and the steps will be discussed soon. We summarize the proposed simultaneously image registration and fusion (SIRF) for pan-sharpening in Algorithm \ref{alg:JTVFusion}.

 \begin{algorithm}[H]
\caption{SIRF} \label{alg:JTVFusion}
\begin{algorithmic}
\STATE {\bfseries Input:} $L$, $\lambda$, $t^{1}=1$, $\textbf Y^0$
\FOR{$k=1$ {\bfseries to} $Maxiteration$}
\STATE $\textbf Y=\textbf Y^k- \psi^T (\psi \textbf{X} -
\textbf{M})/L$
\STATE $\textbf X^k=\arg\min_{\textbf
X}\{\frac{L}{2}\|\textbf X-\textbf Y\|_F^2  +
\lambda||\nabla{\textbf{X}} -\nabla \mathcal{T}({D}(\textbf{P})) ||_{2,1} \}$
\hfill (\textbf{Algorithm} \ref{alg:fjgp})
\STATE $\mathcal{T}=\arg\min_{\mathcal{T}} \{E(\mathcal{T}) = ||\nabla{\textbf{X}}^k - \nabla \mathcal{T}({D}(\textbf{P}))||_{2,1}$\}\\ \hfill (\textbf{Algorithm} \ref{alg:gd})
\STATE $ t^{k+1}=[1+\sqrt{1+4(t^{k})^{2}}]/2$ \STATE $ \textbf
Y^{k+1}=\textbf{X}^{k}+\frac{t^{k}-1}{t^{k+1}}(\textbf X^{k}-\textbf
X^{k-1})$ \ENDFOR
\end{algorithmic}
\end{algorithm}

Here $\psi^T$ denotes the inverse operator of $\psi$. $L$ is the
Lipschitz constant for $\psi^T (\psi \textbf{X} - \textbf{M})$.
We could observe that the
solution is updated based on both $\textbf{X}^k$ and $\textbf{X}^{k-1}$, while the Bregman method that used in previous methods
\cite{ballester2006variational,fang2013variational} updates $\textbf{X}$ only based on $\textbf{X}^k$. This is a reason why our method converges faster.

There are two subproblems in SIRF. For the first subproblem, $L=1$ and
\begin{eqnarray}
\textbf{X}^k = \arg\min_{\textbf{X}}\{\frac{1}{2}\|\textbf{X}-\textbf{Y}\|_F^2+\lambda||\nabla{\textbf{X}} - \nabla \mathcal{T}({D}(\textbf{P}))||_{2,1} \}.
\label{eqn:xdenoising}
\end{eqnarray}
Let $\textbf{Z} = \textbf{X} - \mathcal{T}(D(\textbf{P}))$ and we can rewrite the problem:
\begin{eqnarray}
\textbf{Z}^k= \arg\min_{\textbf{Z}}\{\frac{1}{2}\|\textbf{Z}-(\textbf{Y} - \mathcal{T}(D(\textbf{P}) ))\|_F^2+\lambda||\nabla{\textbf{Z}}||_{2,1} \}.
\label{eqn:zdenoising}
\end{eqnarray}
This alternative problem is therefore a VTV denoising problem
\cite{bresson2008fast,huang2012fast,chen2013calibrationless} and $\textbf{X}^k$ can be updated by
$\textbf{Z}^k + \mathcal{T}(D(\textbf{P}))$. The slow version of the VTV denoising
algorithm \cite{bresson2008fast} is accelerated based on FISTA
framework to solve (\ref{eqn:xdenoising}), which is summarized in
Algorithm \ref{alg:fjgp}. 




The linear operator is defined as:
$\mathcal{L}(\textbf{R},\textbf{S})_{i,j,d}=\textbf{R}_{i,j,d}-\textbf{R}_{i-1,j,d}+\textbf{S}_{i,j,d}-\textbf{S}_{i,j-1,d}$
The corresponding inverse operator is defined as
$\mathcal{L}^{T}(\textbf{X})=(\textbf{R}, \textbf{S})$ with
$\textbf{R}_{i,j,d} = \textbf X_{i,j,d}-\textbf X_{i+1,j,d}$ and $\textbf S_{i,j,d} =
\textbf X_{i,j,d}-\textbf X_{i,j+1,d}$.
$\mathbb{P}$ is a projection operator used to ensure that $\sum_{d=1}^{s}(\textbf{R}_{i,j,d}^2+\textbf{S}_{i,j,d}^2)\leq 1$,
$|\textbf{R}_{i,n,d}|\leq 1$, and $|\textbf{S}_{m,j,d}|\leq 1$. More details can be found in \cite{huang2012fast,chen2013calibrationless}.

\begin{algorithm}[H]
\caption{VTV-Denoising} \label{alg:fjgp}
\begin{algorithmic}
\STATE {\bfseries Input:} $\lambda$, $\textbf{Y}$, $\textbf{P}$, $\mathcal{T}$,
$(\textbf U, \textbf V)=(\textbf R, \textbf S)=(\mathbf{0},
\mathbf{0})$, $t^1 = 1$
\STATE $\textbf{B} = \textbf{Y} - \mathcal{T}(D(\textbf{P}))$
\FOR{$k=1$ {\bfseries to} $Maxiteration$} \STATE$
(\textbf R^{k},\textbf S^{k})=\mathbb{P}[(\textbf U^k,\textbf
V^k)+\frac{1}{8\lambda}\mathcal{L}^{T}(\textbf B-\lambda
\mathcal{L}(\textbf U^k,\textbf V^k))]$
\STATE$ t^{k+1}=\frac{1+\sqrt{1+4(t^{k})^{2}}}{2}$
\STATE$ (\textbf U^{k+1},\textbf V^{k+1})=(\textbf R^{k},\textbf
S^{k})+\frac{t^{k}-1}{t^{k+1}}(\textbf R^{k}-\textbf R^{k-1},\textbf
S^{k}-\textbf S^{k-1})$
\ENDFOR
\STATE $\textbf Z=\textbf B-\lambda\mathcal{L}(\textbf R^k,\textbf S^k)$
\STATE $\textbf{X} = \textbf{Z} + \mathcal{T}(D(\textbf{P}))$
\end{algorithmic}
\end{algorithm}

We solve the second subproblem (\ref{eqn:sub2}) with the gradient descent method with backtracking. One of the advantages is that no extra parameters are introduced during the minimization.

The $\ell_{2,1}$ norm is not smooth.
Let $\textbf{r} = \nabla{\textbf{X}} - \nabla \mathcal{T}({D}(\textbf{P}))$ be the residual and we can have a tight approximation for the function:


\begin{eqnarray}
 E(\mathcal{T}) \approx \sum_{i}\sum_{j}  \sqrt{\sum_{d} (\nabla_{1}\textbf{r}_{i,j,d})^2+(\nabla_{2}\textbf{r}_{i,j,d})^2 + \epsilon}, \label{eqn:DTVappro}
\end{eqnarray}
where $\epsilon$ is a small constant (\emph{e.g.}, $10^{-10}$).

Now, it is not hard to obtain the gradient of the energy function by the chain rule:
\begin{eqnarray}
 \nabla E(\mathcal{T}) = - \frac{\partial E(\mathcal{T})}{\partial \textbf{r}} \nabla \mathcal{T}({D}(\textbf{P})) \frac{\partial \mathcal{T}}{\partial \theta},
  \label{eqn:DTVgrad}
\end{eqnarray}
where 
$\nabla \mathcal{T}({D}(\textbf{P}))$ denotes the image (intensity) gradient;
$\theta$ denotes the parameters of transformation $\mathcal{T}$. 
In this work, the transformation is assumed to be affine or translational, with 6 and 2 motion parameters, respectively.

\begin{figure*}[htbp]
\centering
        \includegraphics[scale=0.46]{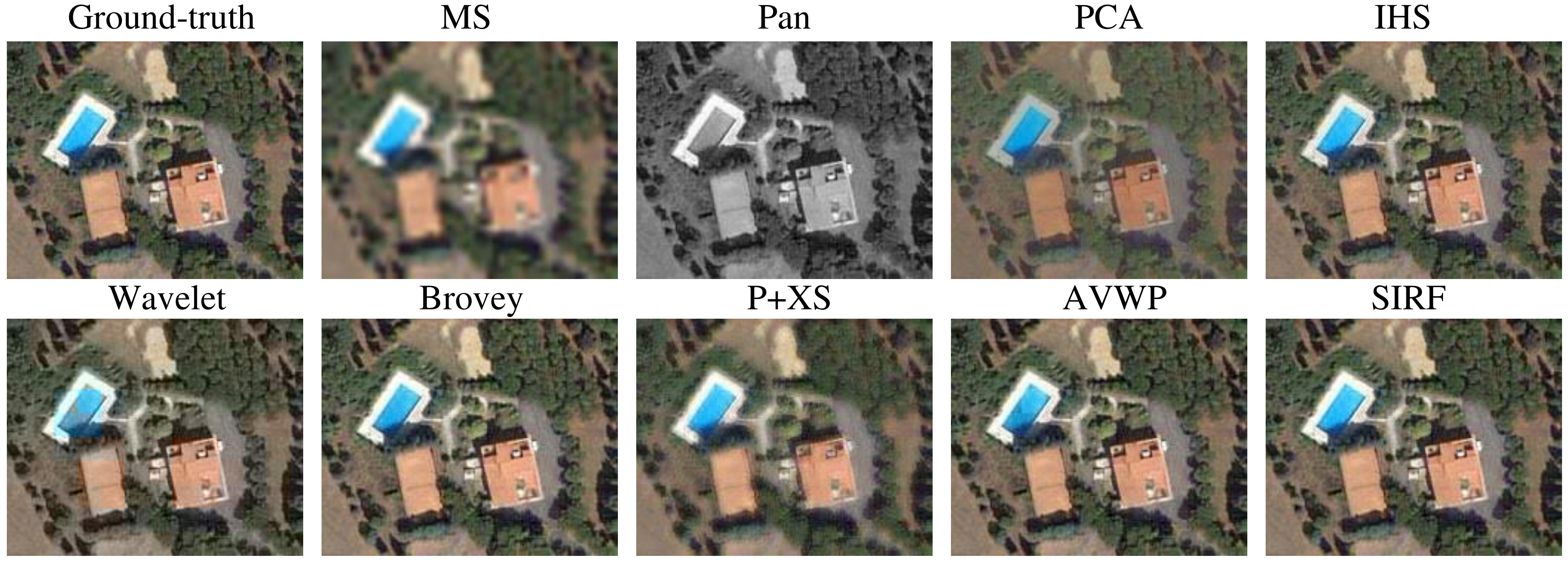}
\caption{Fusion Results comparison (source: Quickbird). The Pan image has $200 \times 160$ pixels. 
}
\label{fig:v1}
\end{figure*}

Gradient descent with backtracking is used to minimize the energy function (\ref{eqn:DTVappro}), which is summarized in Algorithm \ref{alg:gd}. ${\partial \mathcal{T}}/{\partial \theta}$ is calculated based on first order Tyler approximation. We set the initial step size $t^0 = 1$ and $\eta = 0.8$.  It can be easily observed that all operations are linear. Hierarchial estimation is applied for the registration \cite{bergen1992hierarchical}.
The function value is calculated on the overlapped area of two images. To avoid trivial solutions such as zooming in on a dark region,
we use the normalized function value here (divided by the overlapped pixels $M$). When there is no overlapping, the function value will be infinity. We found this approach could effectively rule out the trivial solutions.

 \begin{algorithm}[H]
\caption{Gradient descent with backtracking} \label{alg:gd}
\begin{algorithmic}
   \STATE {\bfseries Input:}  $\textbf{X}, {{D}(\textbf{P})}$, $t^0$, $\eta<1$, $\mathcal{T}^0$, $k=0$.
   \REPEAT
   \STATE $1) \quad$  {compute} $\mathcal{T}^{k+1} = \mathcal{T}^{k} - t^k \nabla E(\mathcal{T}^{k})$
   \STATE $2) \quad$  {if} $E(\mathcal{T}^{k+1})/M_{k+1} > E(\mathcal{T}^{k})/M_k$, {set} $t^k = \eta t^k$ and go back to (1)
   \STATE $3) \quad      t^{k+1} = t^{k}$
   \STATE $4) \quad      k = k+1$
   \UNTIL{Stop criterions}
\end{algorithmic}
\end{algorithm}


Due to the tradeoff between
accuracy and computational cost for solving the subproblem, the inner loops of Algorithms
\ref{alg:fjgp} and \ref{alg:gd} only run three iterations in all experiments.
It has been shown that such inexact solution in FISTA does not change the overall convergence rate \cite{schmidt2011convergence}.


\vspace{-0.0cm}
\begin{figure*}[htbp]
\centering \vspace{-0.0cm}
        \includegraphics[scale=0.365]{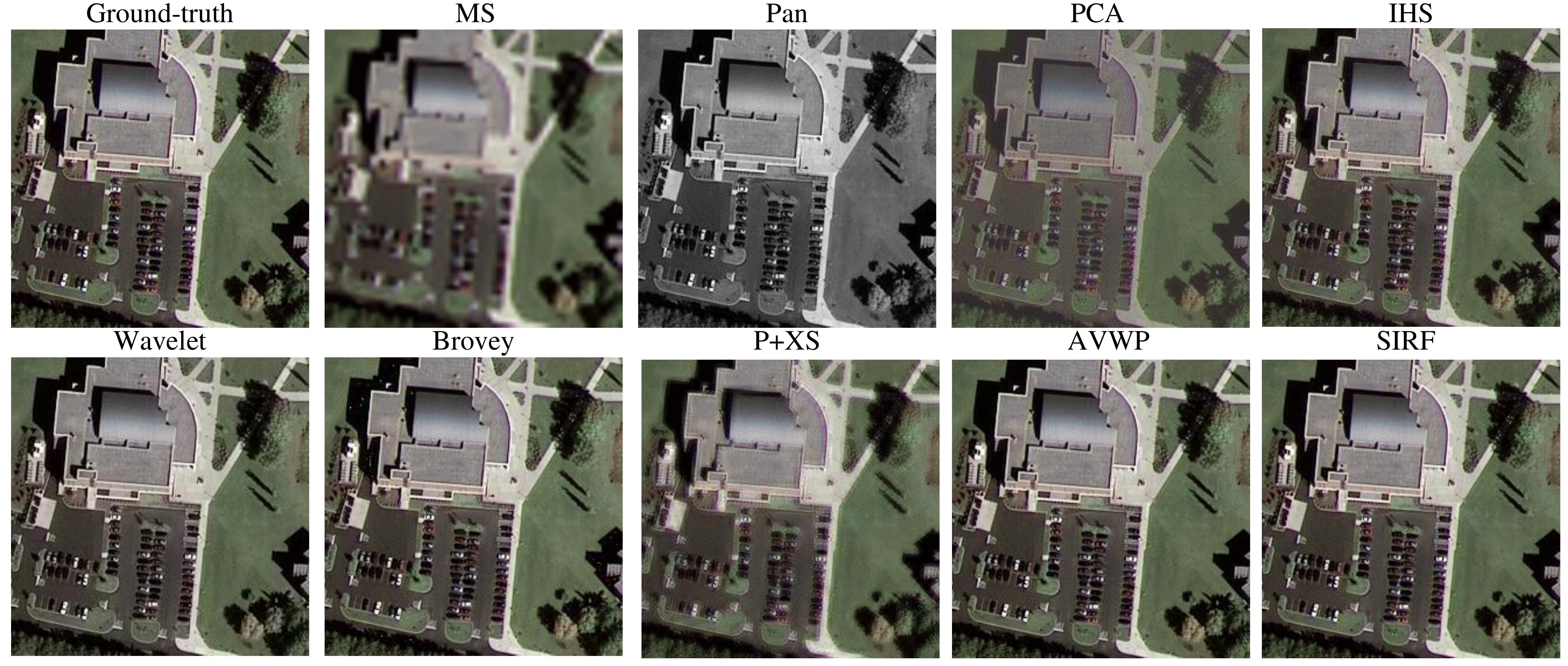}
\caption{Fusion Results comparison (source: IKONOS). The Pan image has $256 \times 256$ pixels.}
\label{fig:v2}\vspace{-0.0cm}
\end{figure*} \vspace{-0.0cm}
\section{Experiments} \label{sec:exp}

\vspace{-0.0cm}
\begin{figure*}[htbp]
\centering \vspace{-0.0cm}
        \includegraphics[scale=0.34]{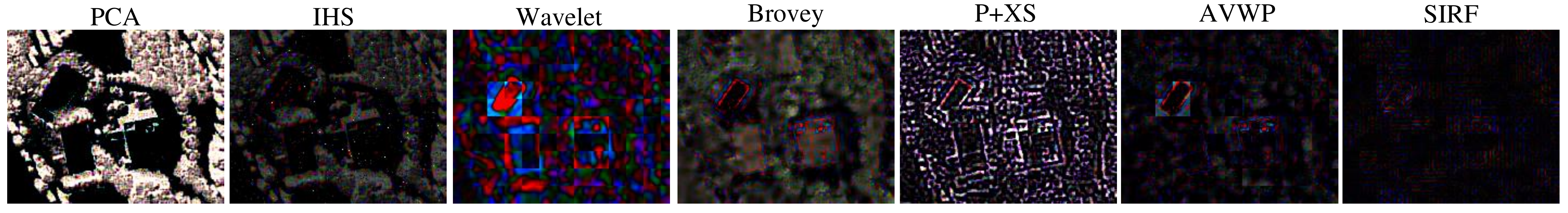}
\caption{The corresponding error images to those in Figure \ref{fig:v1}.
Brighter pixels represent larger errors.}
\label{fig:e1}\vspace{-0.0cm}
\end{figure*} \vspace{-0.0cm}

\vspace{-0.0cm}
\begin{figure*}[htbp]
\centering \vspace{-0.0cm}
        \includegraphics[scale=0.26]{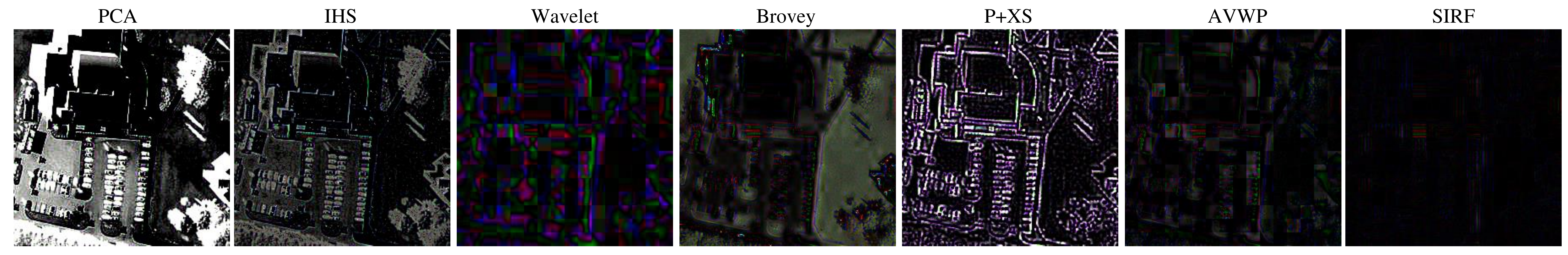}
\caption{The corresponding error images to those in Figure \ref{fig:v2}.
Brighter pixels represent larger errors.}
\label{fig:e2}\vspace{-0.0cm}
\end{figure*} \vspace{-0.0cm}

\subsection{Simulation}

The proposed method is validated on multispectral datasets from Quickbird, Geoeye,
SPOT and IKONOS satellites. The resolution of Pan images ranges from
0.41 m to 1.5 m.  All the corresponding MS images have lower
resolutions with $c=4$ and contain blue, green, red and
near-infrared bands. For convenience, only the RGB bands
are presented.
Due to the lack of multi-resolution images of the
same scene, the original images are viewed as ground-truth and
low-resolution images are downsampled from the
ground-truth images. This strategy is common for comparing fusion
algorithms (\emph{e.g.},
\cite{vijayaraj2004quality,wald1997fusion,ballester2006variational,moller2012variational,fang2013variational}).
In these simulations, there is no misalignment for the input images.
We do not run the registration steps in SIRF except the case with artificial translations.


We compare our method with classical methods PCA
\cite{chavez1991comparison}, IHS \cite{haydn1982application},
wavelet \cite{zhou1998wavelet}, Brovey \cite{gillespie1986color} and
variation methods P+XS \cite{ballester2006variational}, AVWP
\cite{moller2012variational}. 
The recent model-based fusion method (MBF) \cite{aly2014regularized} requires multiple parameters to be tuned based on the individual satellite. The current version can only support the datasets from the IKONOS satellite. Therefore, it is not compared in these simulations.
The effectiveness of our method is validated via extensive experiments with visual and quantitative analysis. Comparisons with P+XS \cite{ballester2006variational} and AVWP \cite{moller2012variational} demonstrate its efficiency.
The parameters for each method are tuned individually
according to the authors' suggestions and the best set is selected
for each method, respectively. 
All experiments are conducted using
MATLAB on a desktop computer with 3.4GHz Intel core i7 3770 CPU\footnote{The demo code of SIRF can be downloaded from: \url{https://dl.dropboxusercontent.com/u/58080329/codeSIRF.zip}}. The simulation results are shown in Subsections \ref{sec:subv} to \ref{sec:subt}.

\subsection{Visual Comparison}  \label{sec:subv}

First, we compare the fusion results by our method with those of
previous works
\cite{chavez1991comparison,haydn1982application,zhou1998wavelet,gillespie1986color,ballester2006variational,moller2012variational}.
Figure \ref{fig:v1} shows the fusion results as well as the ground-truth Quickbird images. All the methods can produce
images with higher resolutions than the original MS image. Obviously, PCA
\cite{chavez1991comparison} performs the worst as the overall intensities of the image has been changed. No obvious artifacts can be
found on the images produced by IHS \cite{haydn1982application} and
Brovey \cite{gillespie1986color}. However, a closer look shows that
the color on these images tends to change, especially on the trees
and grass. This is a sign of spectral distortion
\cite{thomas2008synthesis}. Wavelet fusion \cite{zhou1998wavelet}
suffers from both spectral distortion and  blocky artifacts (\emph{e.g.}, on
the swimming poor). Blurred edges is a general issue in the image
fused by P+XS \cite{ballester2006variational}. AVWP
\cite{moller2012variational} performs much better than all of them
but it inherits the blocky artifacts of the wavelet fusion. The
results of another experiment on a IKONOS image are shown in Figure
\ref{fig:v2}, with similar performance by each algorithm. Some visible bright pixels can be found at the top-left corner
of Brovey.

For better visualization, the error images compared with the
ground-truth are presented in Figure \ref{fig:e1} and Figure \ref{fig:e2}
at the same scale. From these error images, the spectral distortion,
blocky artifacts, and blurriness can be clearly observed. These results
are consistent with those presented in previous work
\cite{moller2012variational}. Due to the spectral distortion, the
conventional methods are not adapted to vegetation study
\cite{thomas2008synthesis}. Previous variational methods
\cite{ballester2006variational,moller2012variational} try to
break such hard assumptions by combining a few weak assumptions. However, such combinations involves more parameters that required to be tuned.
Moreover, the fusion from the upsampled MS image often results in inaccuracy.
In contrast, we only
constrain the spectral information of the fused image to be locally
consistent with the original MS image. The fusion results are
impressively good on these two images.   



\begin{table*}[htbp]
\caption{Performance Comparison on the 158 remotely sensed images.}\label{table:res}
\centering
\begin{tabular}{c|cccccccc}
\hline \hline
Method &ERGAS  &QAVE &RASE  &SAM & FCC &PSNR &MSSIM  &RMSE \\
\hline
PCA \cite{chavez1991comparison}       &5.67$\pm$1.77    &0.664$\pm$0.055        &22.3$\pm$6.8   &2.11$\pm$1.35  &0.972$\pm$0.014          &20.7$\pm$2.7 &0.799$\pm$0.067 &24.1$\pm$6.7\\
IHS \cite{haydn1982application}       &1.68$\pm$0.86    &0.734$\pm$0.011        &6.63$\pm$3.4   &0.79$\pm$0.54  &0.989$\pm$0.006          &31.2$\pm$4.6 &0.960$\pm$0.035 &8.1$\pm$4.2\\
Wavelet\cite{zhou1998wavelet}         &1.18$\pm$0.45    &0.598$\pm$0.113        &4.50$\pm$1.6   &2.45$\pm$1.18  &\textbf{0.997$\pm$0.002} &36.1$\pm$3.6 &0.983$\pm$0.009 &4.5$\pm$1.9\\
Brovey \cite{gillespie1986color}      &1.22$\pm$1.08    &0.733$\pm$0.011        &5.18$\pm$4.6   &0.61$\pm$0.58  &0.940$\pm$0.170          &38.2$\pm$5.6 &0.989$\pm$0.008 &9.1$\pm$19.7\\
P+XS\cite{ballester2006variational}   &0.89$\pm$0.33    &0.720$\pm$0.036        &3.47$\pm$1.3   &0.66$\pm$0.36  &0.898$\pm$0.024          &25.9$\pm$3.5 &0.854$\pm$0.051 &14.7$\pm$5.4\\
AVWP\cite{moller2012variational}      &0.46$\pm$0.17    &0.733$\pm$0.013        &1.81$\pm$0.6   &0.69$\pm$0.70  &0.996$\pm$0.002          &40.0$\pm$3.5 &0.991$\pm$0.006 &2.9$\pm$1.0\\
SIRF   &\textbf{0.07$\pm$0.03}  &\textbf{0.746$\pm$0.004}   &\textbf{0.3$\pm$0.1}   &\textbf{0.18$\pm$ 0.11}    &\textbf{0.997$\pm$0.002} &\textbf{47.5$\pm$3.6} &\textbf{0.998$\pm$0.001} &\textbf{1.1$\pm$0.5} \\
\hline
Desired Value &0 &1 &0 &0  &1 &$+\infty$ &1 &0 \\
\hline \hline
\end{tabular}
\end{table*}

\subsection{Quantitative Analysis}\label{sec:subq}

\begin{figure}[htbp]
\centering \vspace{-0.0cm}
        \includegraphics[scale=0.15]{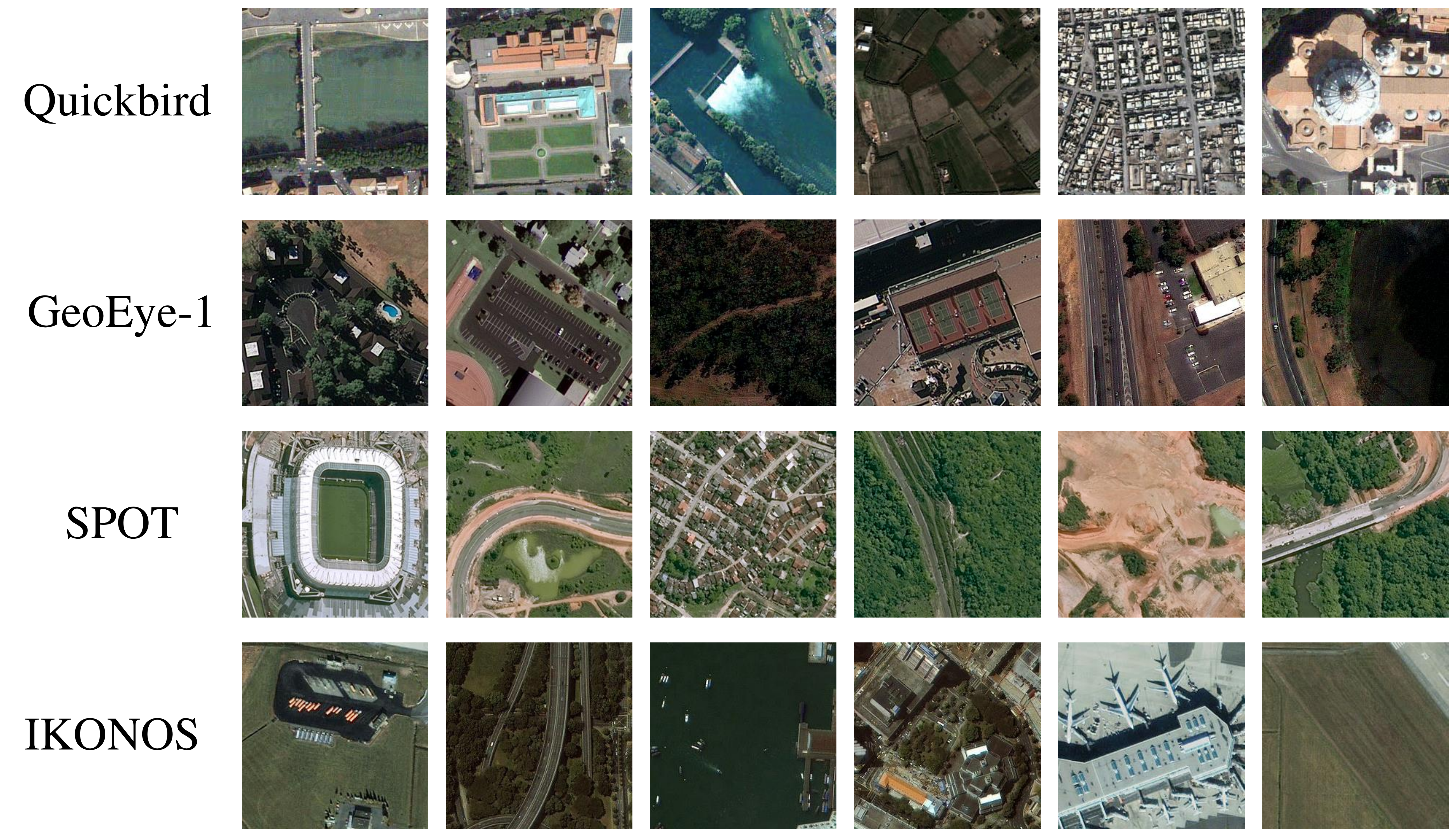}
\caption{Example images used in our experiments. 
} \label{fig:exam}\vspace{-0.0cm}
\end{figure}

In addition to the two images used previously, 156 test images of
different sizes (from $128\times128$ to $512\times512$) are cropped
from Quickbird, Geoeye, IKONOS and SPOT datasets, which contain
vegetation (\emph{e.g.}, forest, farmland), bodies of water (\emph{e.g.}, river, lake)
and urban scenes (\emph{e.g.}, building, road). This test set is much
larger than the size of all datasets considered in previous
variational methods (31 images in \cite{ballester2006variational}, 7
images in \cite{moller2012variational} and 4 images in
\cite{fang2013variational}). Example images are shown in Figure
\ref{fig:exam}.

To evaluate the fusion quality of different methods, we use four metrics that measure spectral quality and one metric that measures spatial quality. The spectral metrics include the relative dimensionless global error in synthesis (ERGAS) \cite{alparone2007comparison}, spectral angle mapper (SAM) \cite{alparone2007comparison}, universal image quality index (Q-average) \cite{wang2006modern} and relative
average spectral error (RASE) \cite{choi2006new}. The filtered correlation
coefficients (FCC) \cite{zhou1998wavelet} is used as spatial quality metric. In addition, peak signal-to-noise ratio (PSNR), and root mean squared error (RMSE) and mean structural similarity (MSSIM) \cite{wang2006modern}
are used to evaluate the fusion accuracy when compared with the ground-truth.

The average results and the variance on this test set are listed
in Table 2. The ideal value for each metric is shown
in the last row.  The results of variational methods
\cite{ballester2006variational,moller2012variational} have
much lower values in ERGAS and RASE than those of conventional
methods
\cite{chavez1991comparison,haydn1982application,zhou1998wavelet,gillespie1986color}.
From QAVE and SAM, the results are comparable to conventional
methods. We can conclude that these variational methods can preserve
more spectral information. Due to the blurriness, P+XS has the worse
spatial resolution in terms of FCC. In terms of error and similarity metrics
(PSNR, MSSIM, RMSE), AVWP and P+XS are always the second best and
second worst, respectively. Except for the same FCC as the wavelet fusion, our
method is consistently better than all previous methods in terms of all metrics. These results are enough to demonstrate the success of our method, where the
dynamic gradient sparsity can preserve sharp edges and the spectral constraint keeps accurate spectral information.
In terms of PSNR, it can
outperform the second best method AVWP by more than 7 dB.

If we consider the prior information that is used, the performance of
each algorithm is easy to explain. Conventional
projection-substitution methods only treat the input images as vectorial information (\emph{i.e.}, 1D). The difference is the substitution performed
on various projection spaces. However, 2D information such as
edges is not utilized. The edge information has been considered
in both variational methods P+XS \cite{ballester2006variational}
and AVWP \cite{moller2012variational}, although their models can not
effectively exploit this prior information. Promising results, especially by
AVWP, have already achieved over conventional methods. By using the
proposed dynamic gradient sparsity, our method has successfully
learned more prior knowledge provided by the Pan image.
Due to the group sparsity
across different bands, our method is less sensitive to noise. These are why our method consistently outperforms the others.


\subsection{Efficiency Comparison} \label{sec:sube}

\begin{figure}[htbp]
\centering \vspace{-0.0cm}
    \subfigure[]{\label{fig:tree1}
        \includegraphics[scale=0.27]{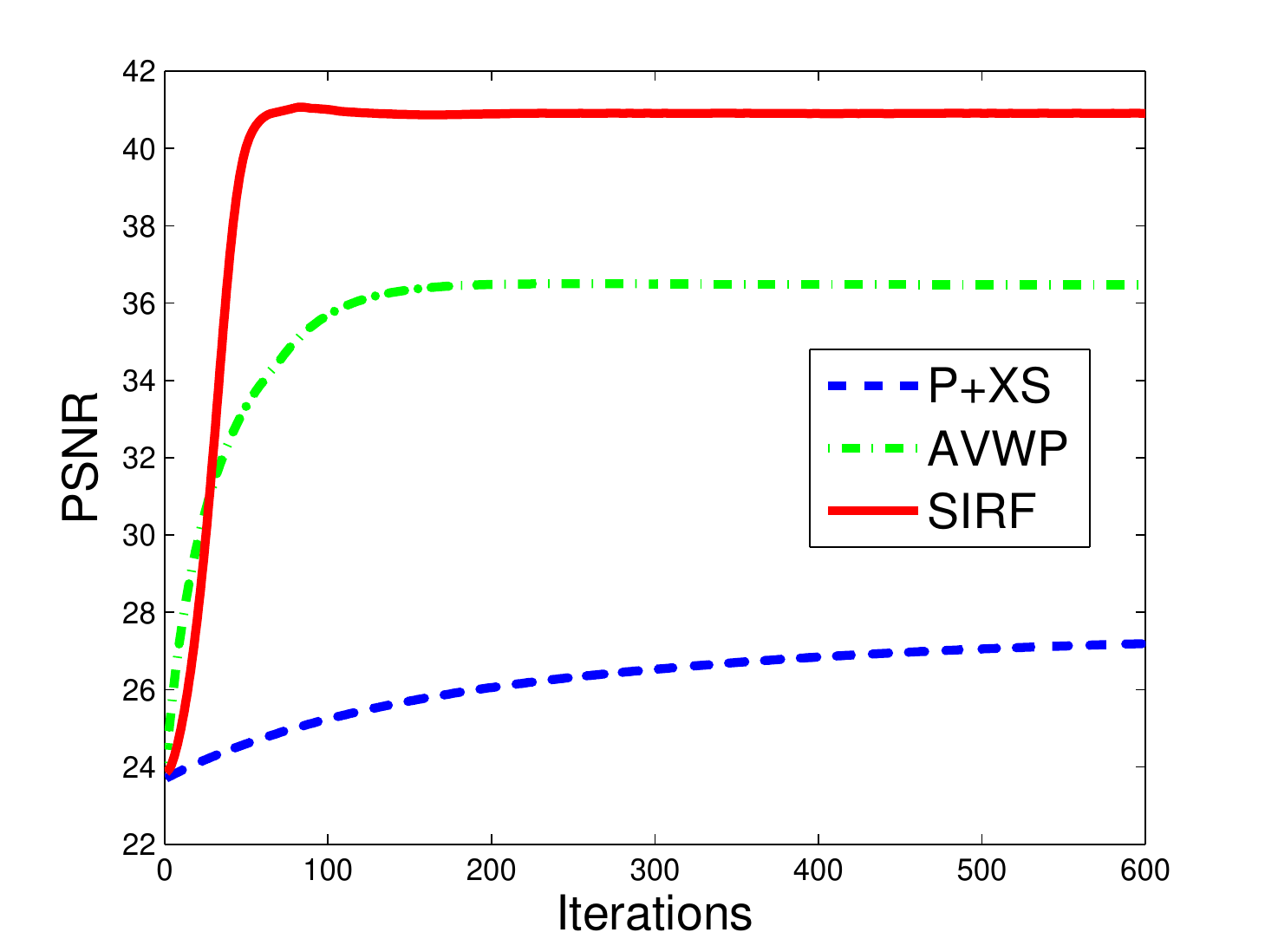}}
    \subfigure[]{\label{fig:brain}
        \includegraphics[scale=0.27]{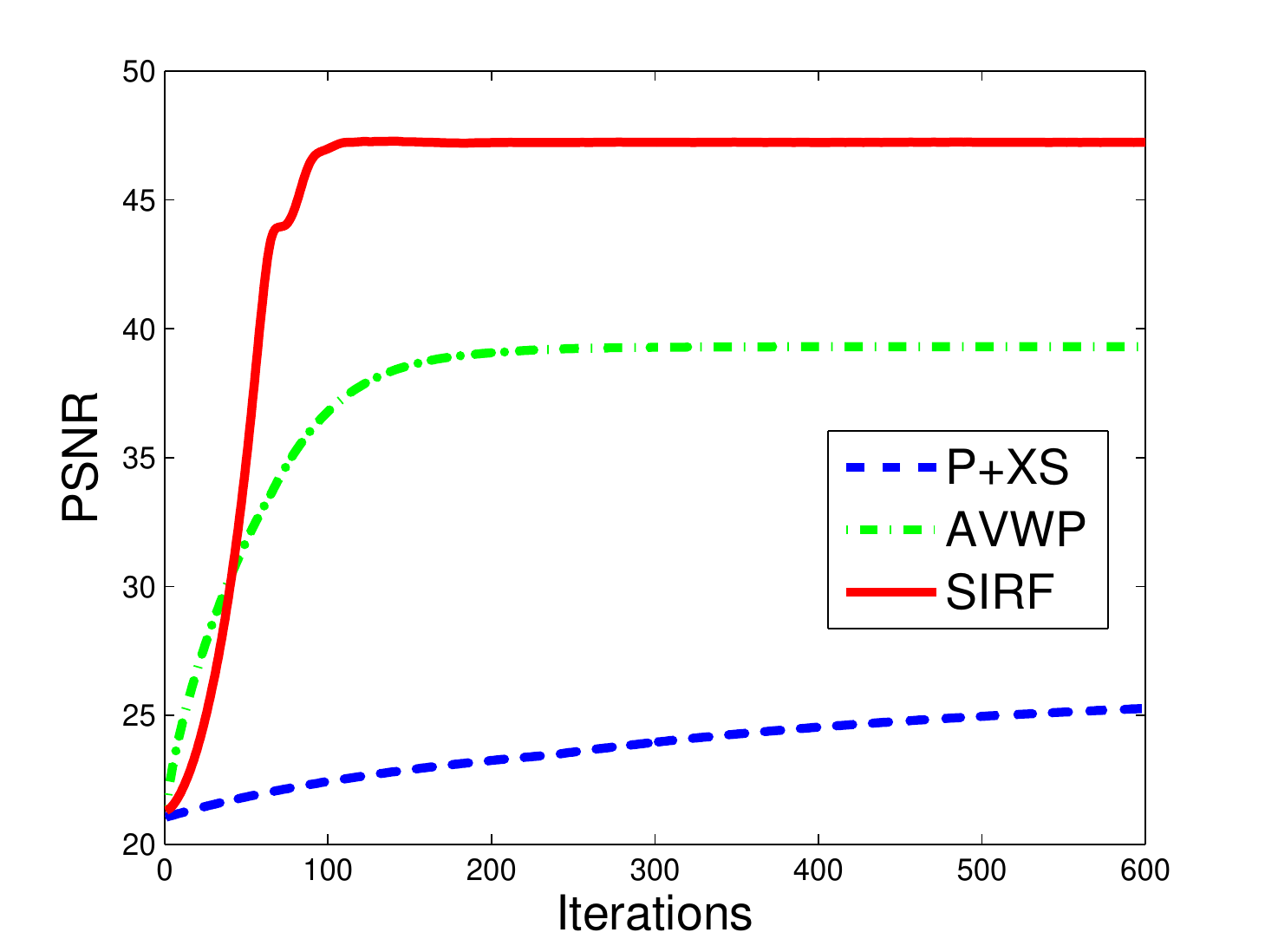}}
\caption{Convergence rate comparison among P+XS, AVWP and the proposed method. (a) Result corresponds to Figure \ref{fig:v1}. (b) Result corresponds to Figure \ref{fig:v2}.}\label{fig:psnr}\vspace{-0.0cm}
\end{figure}

To evaluate the efficiency of the proposed method, we compare the
proposed method with previous variational methods P+XS
\cite{ballester2006variational} and AVWP
\cite{moller2012variational} in terms of both accuracy and
computational cost. PSNR is used to measure fusion accuracy. Figure
\ref{fig:psnr} demonstrates the convergence rate comparison of these
algorithms corresponding to the images in Figure \ref{fig:v1} and
\ref{fig:v2}. Inheriting the benefit of the FISTA \cite{beck2009afast}
framework, our method often converges in 100 to 150 outer
iterations. AVWP often converges in 200 to 400 iterations. P+XS that uses classic gradient decent
method has not converged even with 600 iterations. After each
algorithm converged, our method can approximately outperform AVWP by more than 5 dB and 8 dB on these two images in terms of PSNR. Note that
the later one is the second best method from previous analysis.


The average computational costs of these three methods are listed in Table 3 for different sizes of test images. Both the proposed method and AVWP terminate when a fixed tolerance is reached (\emph{e.g.}, $10^{-3}$ of the relative change on $\textbf{X}$). The computational cost of our method tend to be linear from these results. Even the second fastest method AVWP takes about $50\%$ more time than ours on an image of 512 by 512 pixels.
These comparisons are sufficient to demonstrate the efficiency and effectiveness of our method.

\begin{table}[htbp]
\caption{Computational time (second) comparison.}
\label{table:time}
\begin{tabular}{c|ccccccccc}
\hline
 &$128\times 128$ &$256\times 256$ &$384\times 384$& $512\times 512$\\
\hline
P+XS    &6.7    &16.0 &48.3 &87.4\\
AVWP    &1.7    &8.3 &28.2  &54.7\\
SIRF  &\textbf{1.4} &\textbf{5.0} &\textbf{19.3}    &\textbf{36.8}\\
\hline
\end{tabular}
\end{table}

\subsection{Translation} \label{sec:subt}

\begin{figure}[htbp]
\centering
    \subfigure[]{\label{fig:brain}
        \includegraphics[scale=0.27]{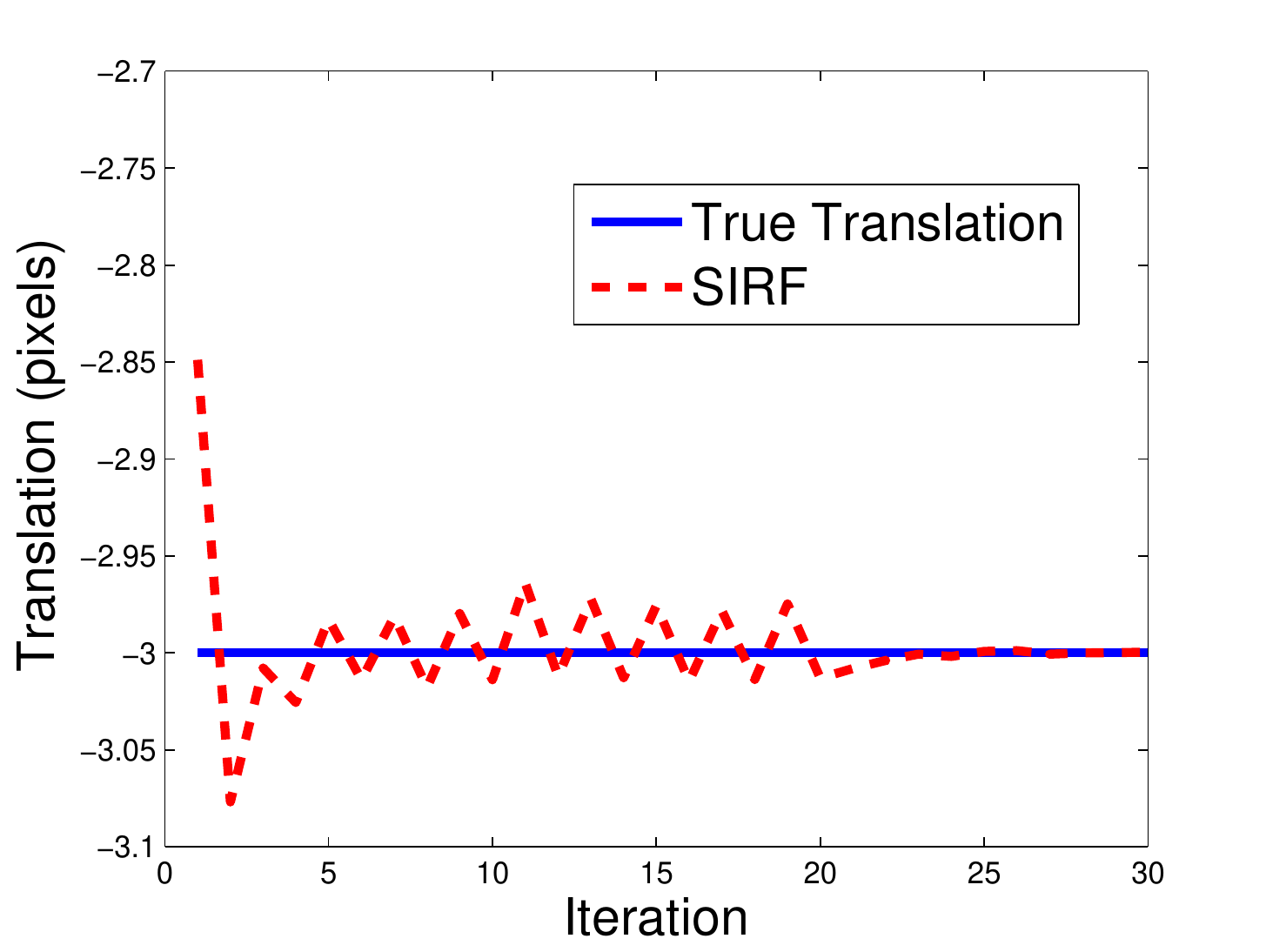}}
    \subfigure[]{\label{fig:tree1}
        \includegraphics[scale=0.27]{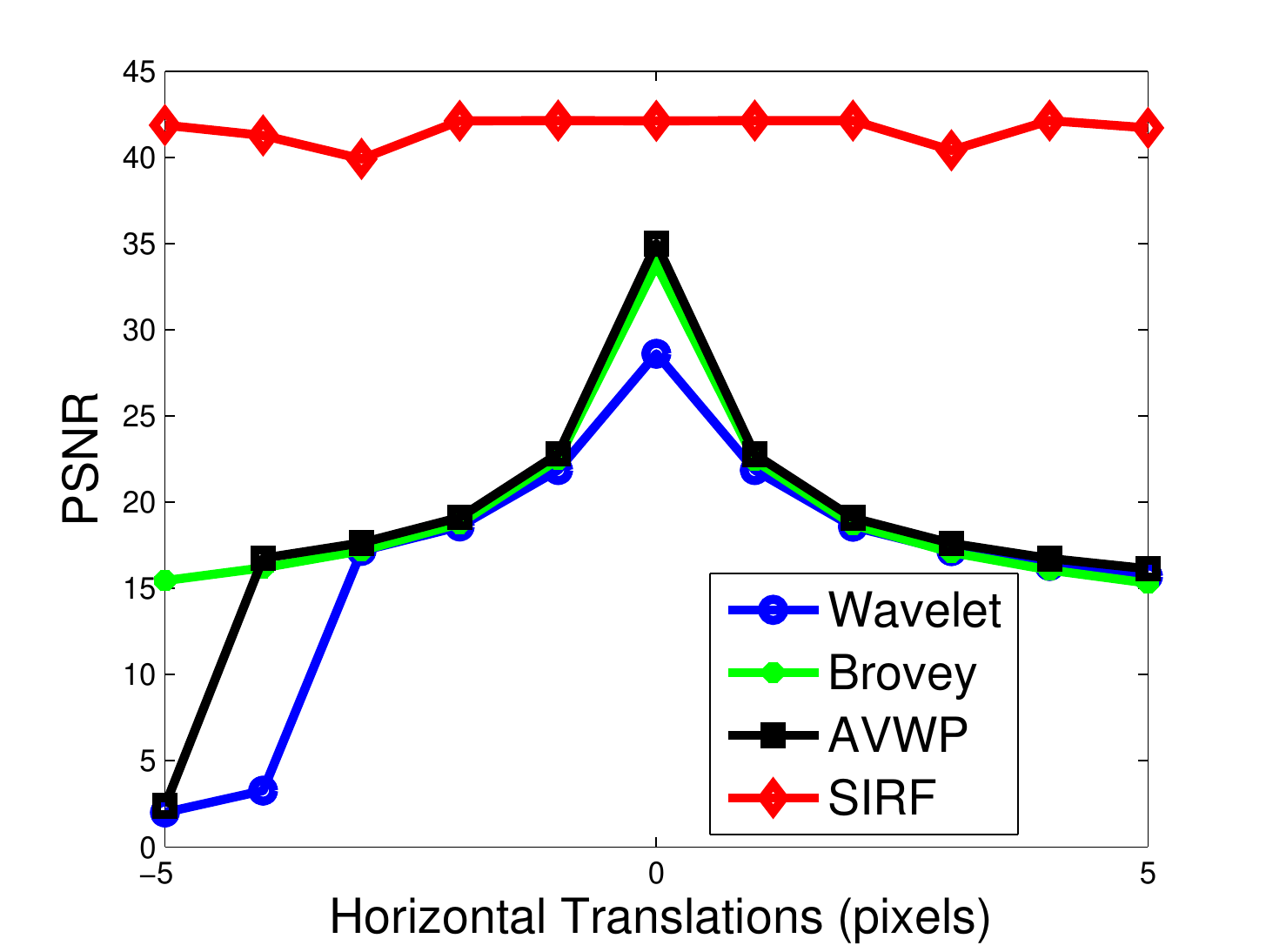}}
\caption{(a) The convergence property of the registration in SIRF. (b) The fusion performance of Wavelet \cite{zhou1998wavelet}, Brovey \cite{gillespie1986color} , AVWP \cite{moller2012variational} and SIRF with respect to horizontal translations on the Quickbird image (Fig. \ref{fig:qb}). }\label{fig:transcomp}
\end{figure}

Since the registration error is almost unavoidable by existing pre-registration approaches, we validate the performance of different methods under various amount of translations.
We add an artificial horizontal translation of 3 pixels in the Pan image shown that in Fig. \ref{fig:qb}. The estimated translation of SIRF is shown in Fig. \ref{fig:transcomp} (a). With round 100 iterations, our method can recover the true translation very accurately. The registration in each iteration costs about 3.7 seconds on this data, which is much slower than the fusion process. Comparing against existing intensity-based registration methods with higher complexity \cite{myronenko2010intensity}, our registration method only has linear complexity to the size of images.
Therefore, such speed is quite acceptable.
Considering this registration cost, we only run the registration in the first 3 iterations in all the later experiments.  With the inexact registration, we find that the error is often below or around 0.03 pixels (\emph{i.e.}, $1\%$ error), and this precision is sufficient for accurate fusion.

\begin{figure}[htbp]
\centering \vspace{-0.0cm}
        \includegraphics[scale=0.50]{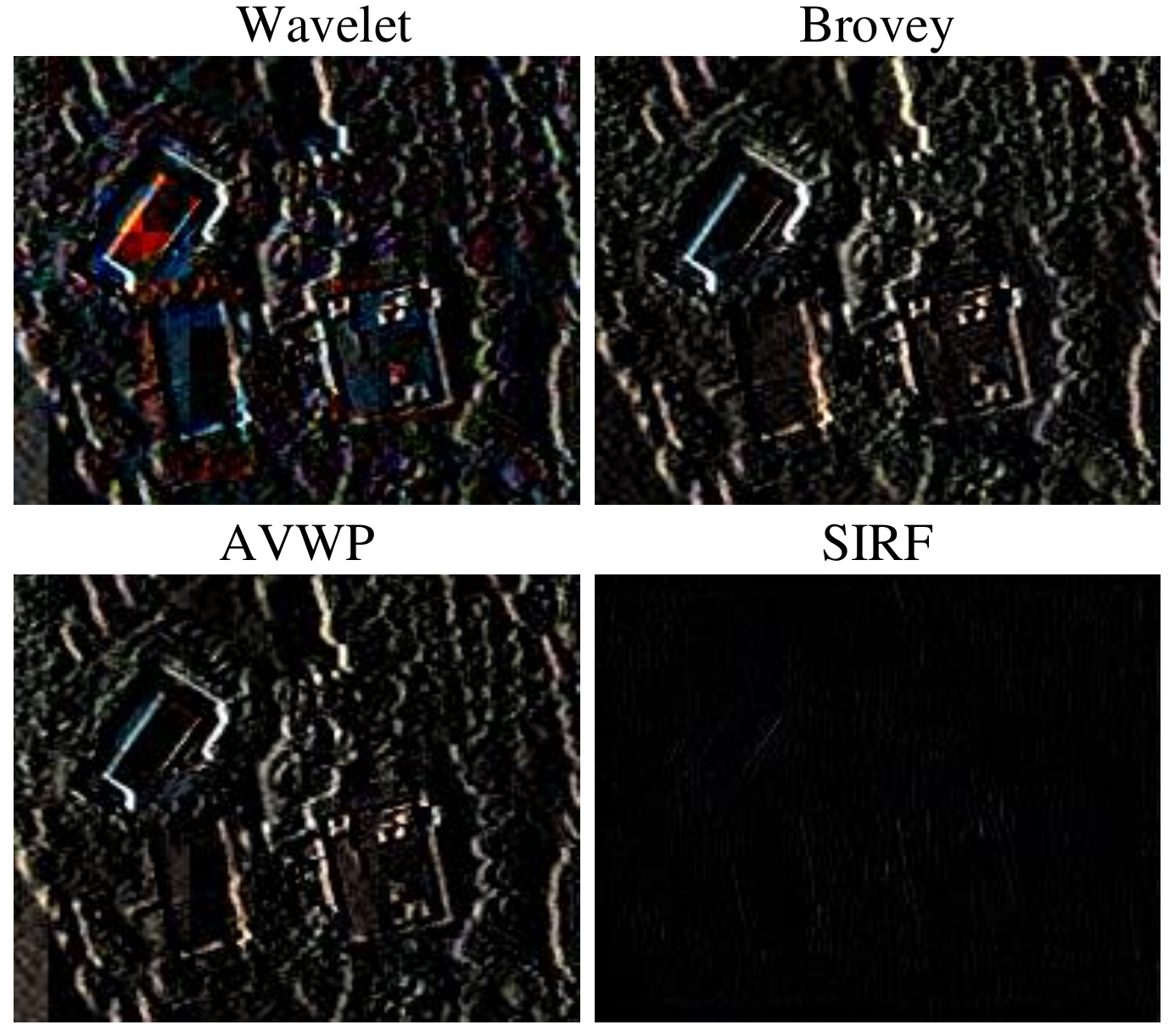}
\caption{The fusion errors of Wavelet \cite{zhou1998wavelet}, Brovey \cite{gillespie1986color} , AVWP \cite{moller2012variational} and SIRF when there is a horizontal misalignment of 3 pixels on the Pan image. The errors are shown at the same scale.}\label{fig:transvisual}\vspace{-0.0cm}
\end{figure}

The best four algorithms in Table \ref{table:res} (based on PSNR), Wavelet \cite{zhou1998wavelet}, Brovey \cite{gillespie1986color}, AVWP \cite{moller2012variational} and SIRF, are selected for comparisons under horizontal translations. The results in Fig. \ref{fig:transcomp} (b) shows that  Wavelet \cite{zhou1998wavelet}, Brovey \cite{gillespie1986color} and AVWP \cite{moller2012variational} cannot "tolerate" large misalignments, while the performance of SIRF is quite stable. The fusion errors are presented in Fig. \ref{fig:transvisual} when there is a horizontal misalignment of 3 pixels on the Pan image. Due to the misalignment, visible errors can be observed on the boundaries of land objects by Wavelet, Brovey and AVWP.
Our method can simultaneously estimate the translation and achieve substantial improvement over existing methods.

\subsection{Real-World Datasets}
\begin{figure*}[htbp]
\centering
        \includegraphics[scale=0.32]{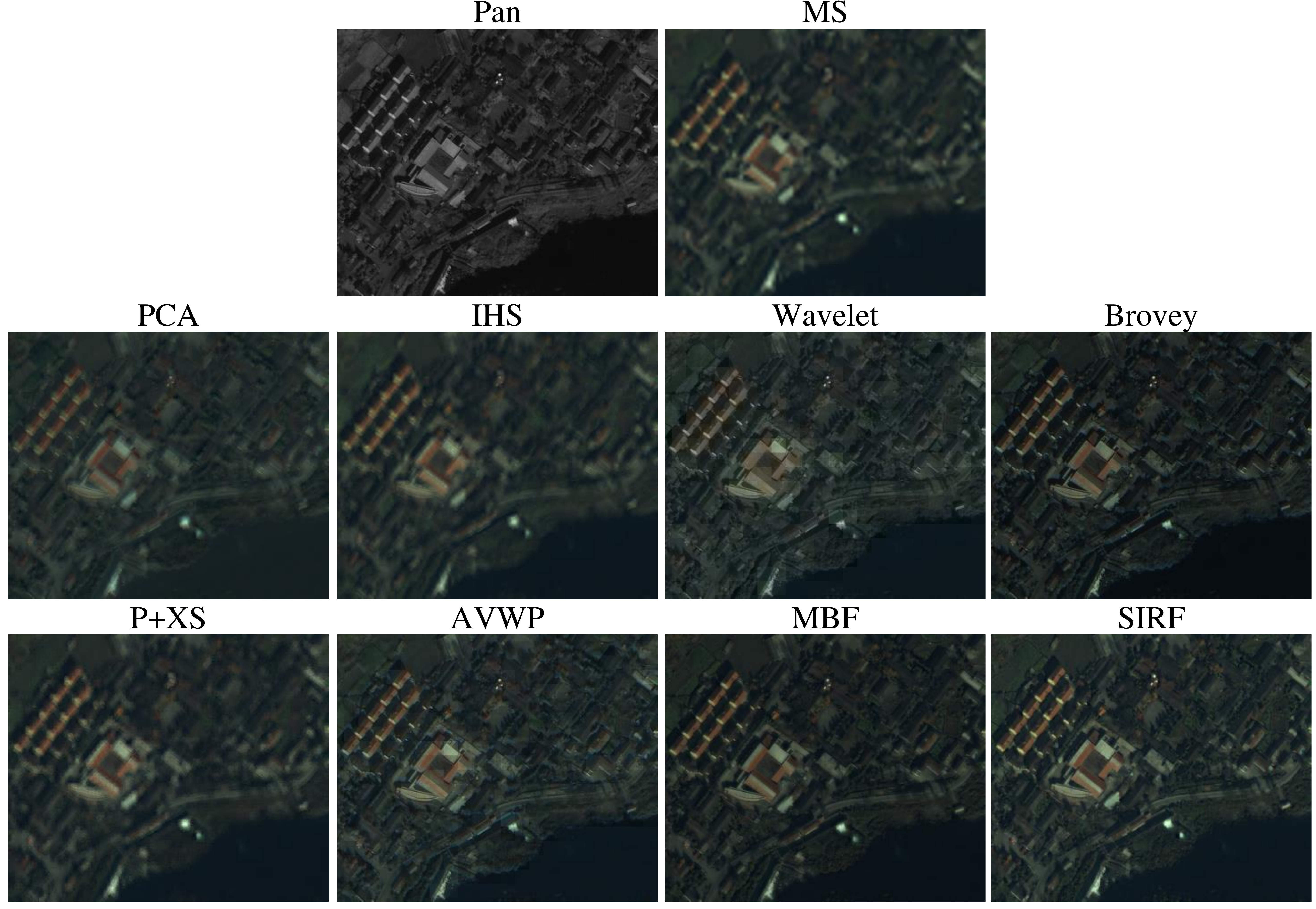}
\caption{The fusion result for a portion of the IKONOS China-Sichuan 58208\_0000000.20001108 dataset. The RGB channels are presented. This figure is better viewed on screen with $200\%$ size.}\label{fig:visualIKONOS}
\end{figure*}

Finally, we evaluate the fusion results of different method on real-world datasets (non-simulated).
The imagery were acquired by IKONOS multispectral imaging satellite \cite{IKONOS}, which contains pre-registered Pan and MS images at their capture resolutions.
The registration error is within a subpixel on the MS images (up to 4 pixels on the Pan image), and we do not add artificial transformation on the datasets.
The multispectral image has four bands: blue, green, red and near-infrared, with 4 meter resolution. The Pan image has 1 meter resolution.
We re-scale these images to 0-255 before processing. The bicubic interpolation is used for both upsampling and downsampling.

In addition to the methods we have compared, the latest method MBF \cite{aly2014regularized} is also compared for these datasets. The parameters of MBF have already been optimized on these datasets and we use their default setting for experiments. Fig. \ref{fig:visualIKONOS} shows the fusion results on a portion of the datasets\footnote{The results are shown on the 0-255 scale without histogram stretch.}. The images obtained by PCA, IHS, and P+XS have blurry edges due to the misalignment. Blocky artifacts can be found in the results by Wavelet and AVWP, which is consistent with the observation in \cite{aly2014regularized}. Brovey, MBF and the proposed SIRF provide high quality images with sharp object boundaries. Such results are reasonable because the pixel-to-pixel based methods are often sensitive to the misalignment. In Brovey, the spectral information is mapped to the Pan image, and the high spatial resolution can always be preserved. Based on the designed high-pass and low-pass filters in MBF, it can somewhat tolerate the misalignment. In SIRF, the transformation is estimated simultaneously during the fusion process. Therefore, accurate fusion can be achieved from the well-aligned images.
\begin{figure}[htbp]
\centering \vspace{-0.0cm}
        \includegraphics[scale=0.19]{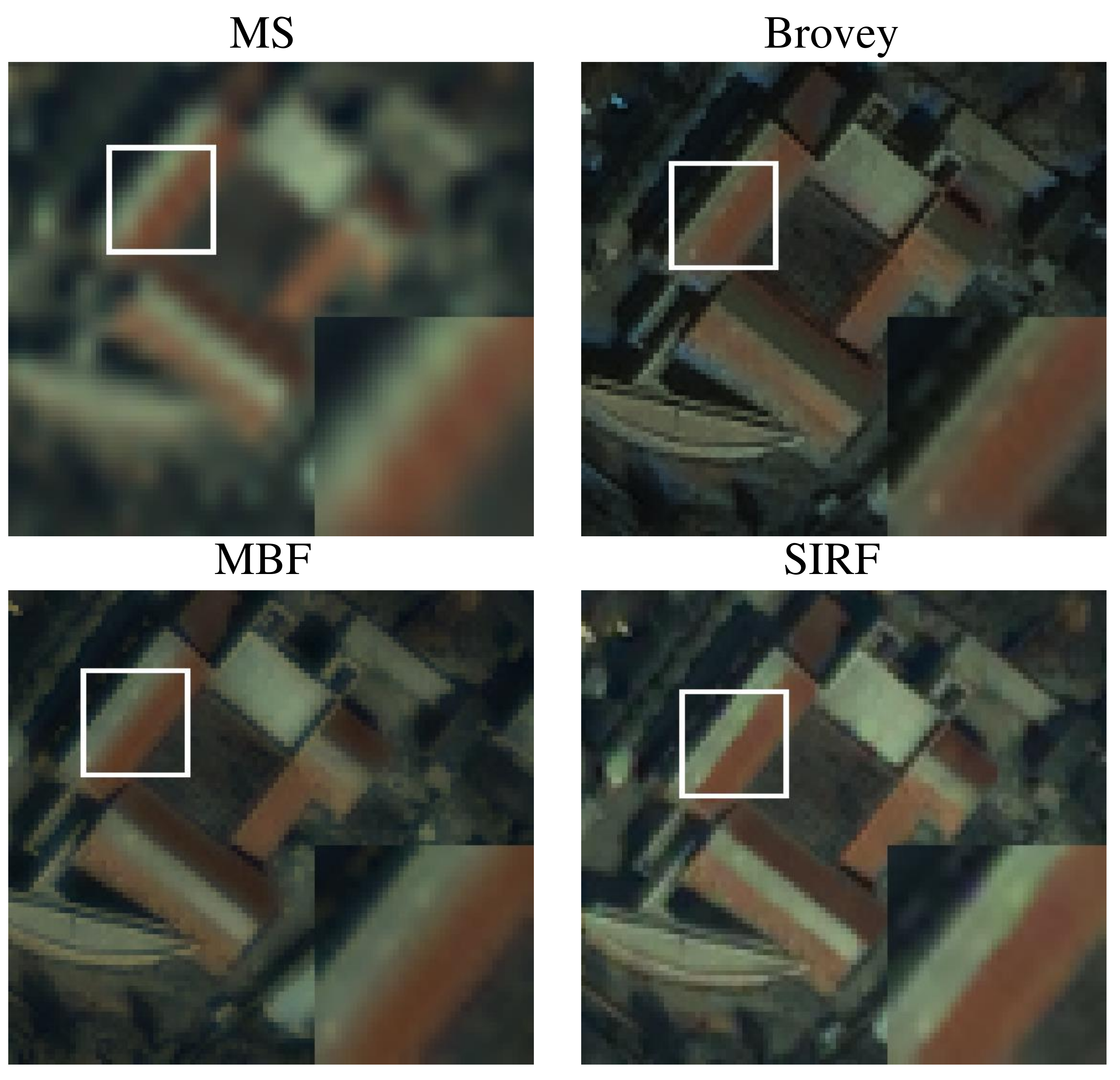}
\caption{The zoomed-in areas fused by Brovey, MBF and SIRF in Fig. \ref{fig:visualIKONOS}.}\label{fig:zoomIKONOS}\vspace{-0.0cm}
\end{figure}

We then take a close look at the fused images by Brovey, MBF and SIRF, which are shown in Fig. \ref{fig:zoomIKONOS}.
From the upsampled MS image, it shows the roofs of the buildings should be in white and dark orange colors. The spectral distortion can be clearly observed in the fused image by Brovey, where the most of the roof is covered by dark orange. Both MBF and SIRF provide results with high quality. The zoomed-in areas show that the result by our method has sharper edges than that by MBF. The two colors are mixed together in the result by MBF, while this mixed color does not actually exist in the original MS image. 
These results are sufficient to demonstrate that the our method outperform the existing methods on this dataset due to the inherent registration scheme.

%


From the previous visual results, MFB and SIRF provide the most accurate results.
We quantitatively compare the results by MFB and SIRF on four portions of the China-Sichuan IKONOS datasets\footnote{We use the same portions as in \cite{aly2014regularized}.}, such as that shown in Fig. \ref{fig:visualIKONOS}. As there is no ground-truth for these real datasets, we only use the ERGAS,  QAVE, RASE, SAM and FCC for comparisons. The results are shown in Table \ref{table:IKONOS}. For both spectral and spatial information, our method is consistently better than MBF. 
Due to the misalignment of pre-registration, the fusion error in MBF is expected. We consider the correlations across different bands and perform the fusion jointly, while MBF
can only fuse the different bands individually. Both the fusion scheme and simultaneous registration contribute to the promising results of SIRF.

\begin{table}[htbp]
\renewcommand{\arraystretch}{1.0}
\caption{Quantitative performance comparison between MBF and SIRF on the IKONOS China-Sichuan datasets.}
\label{table:IKONOS}
\begin{tabular}{|c|ccccc|}
\hline \hline
Method &ERGAS  &QAVE &RASE  &SAM & FCC \\
\hline
Desired Value &0 &1 &0 &0  &1 \\
\hline
\multicolumn{6}{c}{China-Sichuan 58204\_0000000.20001116} \\
\hline
MBF &1.80  &0.973 &7.46 &2.22 &0.887\\
SIRF  &\textbf{0.06} &\textbf{1.000} &\textbf{0.25} &\textbf{0.07} &\textbf{0.919}\\

\hline
\multicolumn{6}{c}{China-Sichuan 58205\_0000000.20001003} \\
\hline
MBF &1.73  &0.980 &7.49 &1.68 &0.861\\
SIRF  &\textbf{0.06} &\textbf{1.000} &\textbf{0.23} &\textbf{0.06} &\textbf{0.920}\\
\hline

\multicolumn{6}{c}{China-Sichuan 58207\_0000000.20000831} \\
\hline
MBF &2.78  &0.947 &11.49 &2.39 &0.793\\
SIRF  &\textbf{0.04} &\textbf{1.000} &\textbf{0.17} &\textbf{0.04} &\textbf{0.869}\\

\hline
\multicolumn{6}{c}{China-Sichuan 58208\_0000000.20001108} \\
\hline
MBF &2.38  &0.954 &9.89 &2.34 &0.915\\
SIRF  &\textbf{0.08} &\textbf{0.999} &\textbf{0.33} &\textbf{0.11} &\textbf{0.927}\\
\hline

\end{tabular}
\end{table}





\section{Conclusion and Discussion} \label{sec:con}

We have proposed a novel and powerful variational model for
simultaneous image registration and fusion in a unified framework, based on the property of dynamic gradient sparsity.
The model naturally incorporates the gradient prior information from a high-resolution Pan image and the spectral information from a low-resolution MS image.
Our method leads to several advantages over existing methods. First, the proposed dynamic gradient sparsity can directly exploit the sharp edges from the Pan image,
which has been shown very effective. Second, we jointly sharpen the images by incorporating the intra-correlations across different bands, while most existing methods
are based on band-by-band fusion. The last and most importantly, although the registration is quite challenging between the Pan and MS images due to their different spatial resolutions, our method can simultaneously register the two input images during the fusing process, acting directly on the input images without any pre-filtering and feature extraction.

An efficient optimization algorithm has been devised to solve the problem, with sufficient implementation details.
Extensive experiments were conducted on 158 simulated images stemming from a variety of sources. Due to the proposed unique
techniques, our method is corroborated to consistently outperform the state-of-the-arts in terms of both spatial and spectral qualities.
We further evaluated our method on real IKONOS datasets with comparison to the latest methods.
The results show that our method can effectively eliminate the misalignment during preregistration, and provide higher quality products than the existing methods.
Due to the high accuracy and simultaneous registration property, our method may benefit more applications in remote sensing, \textit{e.g.}, classification, change detection, \textit{etc}.
The gradient prior information and joint structure have been separately utilized in other image enhancement tasks \cite{krishnan2009dark,liu2013joint}.
Such a success further demonstrates the effectiveness of our modeling.

Parallel programming can further accelerate the running speed of our method. No information from another patch is required for fusion.
As our method does not require a strong correlation between two input images, it may be used in fusing images from different capture times
(shadows should be taken into account) or different sources (\textit{e.g.}, images from different satellites).
Currently, the land objects in the input images are assumed to be consistent. If there exist moving objects (\textit{e.g.}, vehicles),
blurriness may occur at the corresponding locations. Our method may fail in some extreme cases. For example, the input images are acquired by different platforms,
particularly, one image is occluded by clouds while the other one is not. In the future, we will try to cope with such cases.

{
\bibliographystyle{IEEEtran}
\bibliography{IEEEbib_new}
}

\ifCLASSOPTIONcaptionsoff
  \newpage
\fi

\begin{IEEEbiography}[{\includegraphics[width=1in,height=1.25in,clip,keepaspectratio]{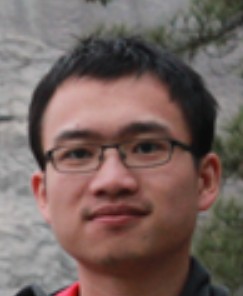}}]{Chen Chen}
received the BE degree and MS degree both from Huazhong University of Science and Technology, Wuhan, China, in 2008 and 2011, respectively.
He has been a PhD student in the Department of Computeter Science and Engineering at the University of Texas at Arlington since 2012.
His major research interests include image processing, medical imaging, computer vision and machine learning. He is a student member of the IEEE.
\end{IEEEbiography}

\begin{IEEEbiography}[{\includegraphics[width=1in,height=1.25in,clip,keepaspectratio]{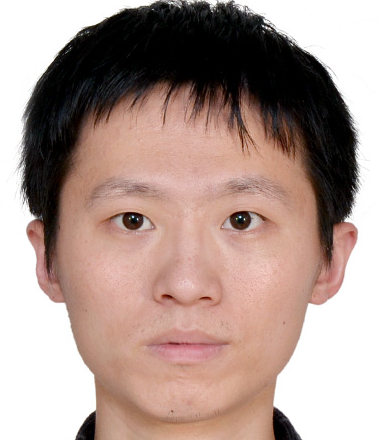}}]{Yeqing Li}
received the BE degree in computer science and technology from Shantou University, China, in 2006, the ME degree
from Nanjing University, Nanjing, China, in 2009 and has been working
toward the PhD degree from the Department of
Computer Science at the University of Texas at Arlington, since 2012. His major
research interests include machine learning,
pattern recognition, medical image analysis,
and computer vision.
\end{IEEEbiography}

\begin{IEEEbiography}[{\includegraphics[width=1in,height=1.25in,clip,keepaspectratio]{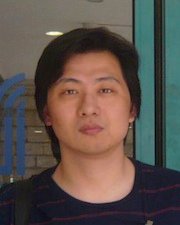}}]{Wei Liu}
received the M.Phil. and Ph.D. degrees in electrical engineering from Columbia University, New York, NY, USA in 2012. Currently, he is a research staff member of IBM T. J. Watson Research Center, Yorktown Heights, NY, USA, and holds an adjunct faculty position at Rensselaer Polytechnic Institute, Troy, NY, USA. He has been the Josef Raviv Memorial Postdoctoral Fellow at IBM T. J. Watson Research Center for one year since 2012. His research interests include machine learning, data mining, computer vision, pattern recognition, and information retrieval. Dr. Liu is the recipient of the 2011-2012 Facebook Fellowship and the 2013 Jury Award for best thesis of Department of Electrical Engineering, Columbia University.
\end{IEEEbiography}

\begin{IEEEbiography}[{\includegraphics[width=1in,height=1.25in,clip,keepaspectratio]{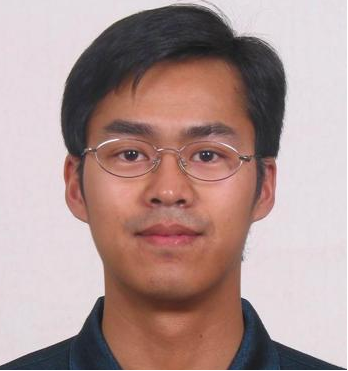}}]{Junzhou Huang}
received the BE degree from
Huazhong University of Science and Technology, Wuhan, China, in 1996, the MS degree
from the Institute of Automation, Chinese Academy of Sciences, Beijing, China, in 2003, and
the PhD degree from Rutgers University, New
Brunswick, New Jersey, in 2011. He is an
assistant professor in the Computer Science
and Engineering Department at the University of
Texas at Arlington. His research interests
include biomedical imaging, machine learning and computer vision,
with focus on the development of sparse modeling, imaging, and
learning for large scale inverse problems. He is a member of the IEEE.
\end{IEEEbiography}




\end{document}